\def\arxivbuild{1}
\ifdefined\arxivbuild
  \documentclass[sigconf,nonacm]{acmart}
\else
  \documentclass[sigconf,anonymous,nonacm]{acmart}
\fi
\usepackage{amsmath,booktabs,tabularx,multirow,tikz}
\newcolumntype{Y}{>{\raggedright\arraybackslash}X}
\setcopyright{none}

\newcommand{\DR}[1]{\textsf{\ifcase#1 Halt\or Contain\or Rules\or Risk-Local\or Risk-Lineage\fi}}
\newcommand{\Ag}[1]{\textsf{\ifcase#1 Halt\or Contain\or Rules\or Risk-Local\or Risk-Lineage\fi+\allowbreak Agent}}
\newcommand{\ValAll}{\textsf{Validate-All}}
\newcommand{\RandMatched}{\textsf{Random-Matched}}

\newcommand{\LlmNumCtx}{8{,}192}
\newcommand{\LlmNumPredict}{512}

\newcommand{\LlmTemperature}{0}

\newcommand{\LlmQuant}{Q4\_K\_M}

\newcommand{\LlmArms}{9}
\newcommand{\LlmEpisodes}{540}

\newcommand{\InfraErrorsAll}{0}

\newcommand{\EpisodesAll}{960}

\newcommand{\ResTwoInv}{37}
\newcommand{\ResTwoRel}{120}

\newcommand{\ResTwoValidators}{0}

\newcommand{\ResTwoRecNum}{12}
\newcommand{\ResTwoRecDen}{12}

\newcommand{\ResThreeRecNum}{2}
\newcommand{\ResThreeRecDen}{2}
\newcommand{\ResThreePTValid}{2}
\newcommand{\ResThreePTFB}{20}

\newcommand{\ResFourInv}{0}

\newcommand{\ResFourRecNum}{4}
\newcommand{\ResFourRecDen}{4}

\newcommand{\ResWfReq}{60}
\newcommand{\DZeroOneValid}{0.267}
\newcommand{\DZeroOneValidCI}{[0.182, 0.346]}

\newcommand{\DOneTwoValid}{0.317}
\newcommand{\DOneTwoValidCI}{[0.180, 0.458]}

\newcommand{\DOneTwoInv}{0.617}
\newcommand{\DOneTwoInvCI}{[0.434, 0.783]}

\newcommand{\DTwoThreeValid}{\ensuremath{-}0.533}
\newcommand{\DTwoThreeValidCI}{[\ensuremath{-}0.952, \ensuremath{-}0.018]}

\newcommand{\DTwoThreeInv}{\ensuremath{-}0.617}

\newcommand{\DTwoFourValid}{0.283}
\newcommand{\DTwoFourValidCI}{[0.073, 0.492]}

\newcommand{\DTwoFourInv}{\ensuremath{-}0.617}
\newcommand{\DTwoFourInvCI}{[\ensuremath{-}0.791, \ensuremath{-}0.433]}

\newcommand{\DTwoFourCost}{8.57}
\newcommand{\DTwoFourCostCI}{[7.56, 9.53]}

\newcommand{\DThreeFourValid}{0.817}
\newcommand{\DThreeFourValidCI}{[0.346, 1.212]}
\newcommand{\DThreeFourValidPT}{1.633}
\newcommand{\DThreeFourValidPTCI}{[1.533, 1.769]}
\newcommand{\DThreeFourValidSIR}{0.000}

\newcommand{\ResInstances}{60}
\newcommand{\ResCaseGroups}{16}
\newcommand{\ResCaseGroupsPT}{6}
\newcommand{\ResCaseGroupsSIR}{10}

\newcommand{\AudLlmCalls}{1{,}020}

\newcommand{\AudLlmErrors}{0}

\newcommand{\RevN}{846}

\newcommand{\RevAgree}{0.844}
\newcommand{\RevAgreeCI}{[0.818, 0.867]}
\newcommand{\RevKappa}{0.00}

\newcommand{\RevDRTwoInvalidApproved}{37}
\newcommand{\RevDRTwoApprovedInvalid}{44}
\newcommand{\RevDRTwoSuperseded}{7}
\newcommand{\GemmaDRTwoTrajDiff}{5}

\newcommand{\DGemmaValid}{\ensuremath{-}0.267}
\newcommand{\DGemmaValidCI}{[\ensuremath{-}0.443, \ensuremath{-}0.117]}

\newcommand{\DGemmaTwoFourValid}{0.017}
\newcommand{\DGemmaTwoFourValidCI}{[\ensuremath{-}0.125, 0.138]}

\newcommand{\CalThreeBrier}{0.0153}
\newcommand{\CalThreeBrierCI}{[0.0108, 0.0218]}
\newcommand{\CalThreeRef}{0.086}

\newcommand{\CalFourBrier}{0.0036}
\newcommand{\CalFourBrierCI}{[0.0015, 0.0076]}
\newcommand{\CalFourRef}{0.085}

\newcommand{\AgentModelsOpen}{3}

\newcommand{\AgentModelsOpenIdentical}{2}

\newcommand{\RiskGatedArmsOpen}{5}
\newcommand{\RiskGatedInvalidOpen}{0}
\newcommand{\LabelAgree}{600}
\newcommand{\LabelTotal}{600}
\newcommand{\AblTwoVInv}{0}
\newcommand{\AblTwoVRel}{119}

\newcommand{\AblTwoVValid}{112}
\newcommand{\AblTwoVReq}{120}

\newcommand{\AblTwoMTwoInv}{3}

\newcommand{\AblTwoMTwoValidators}{251}

\newcommand{\AblFourInsertions}{248}

\newcommand{\DAblFourMTwoInvRate}{\ensuremath{-}0.027}
\newcommand{\DAblFourMTwoInvRateCI}{[\ensuremath{-}0.063, 0.000]}

\newcommand{\DAblFourMTwoYield}{\ensuremath{-}0.008}
\newcommand{\DAblFourMTwoYieldCI}{[\ensuremath{-}0.075, 0.058]}

\newcommand{\DAblFourMTwoCost}{\ensuremath{-}2.12}
\newcommand{\DAblFourMTwoCostCI}{[\ensuremath{-}3.86, \ensuremath{-}0.38]}

\newcommand{\DAblFourVYield}{\ensuremath{-}0.100}
\newcommand{\DAblFourVYieldCI}{[\ensuremath{-}0.167, \ensuremath{-}0.042]}

\newcommand{\DAblFourVCost}{\ensuremath{-}5.52}
\newcommand{\DAblFourVCostCI}{[\ensuremath{-}7.50, \ensuremath{-}3.91]}

\newcommand{\AblFourVsVValPct}{31}
\newcommand{\AblFourVsVCostPct}{22}

\newcommand{\NoHarnLowNum}{5}
\newcommand{\NoHarnLowDen}{120}
\newcommand{\NoHarnLow}{0.042}
\newcommand{\NoHarnLowCI}{[0.008, 0.075]}

\newcommand{\NoHarnUpNum}{72}
\newcommand{\NoHarnUpDen}{120}
\newcommand{\NoHarnUp}{0.600}
\newcommand{\NoHarnUpCI}{[0.492, 0.708]}

\newcommand{\NoHarnPointNum}{55}
\newcommand{\NoHarnPointDen}{120}
\newcommand{\NoHarnPoint}{0.458}
\newcommand{\NoHarnPointCI}{[0.358, 0.558]}

\newcommand{\NoHarnNotProdNum}{67}
\newcommand{\NoHarnNotProdDen}{120}

\newcommand{\NoHarnDRTwoInvSA}{20}

\newcommand{\NoHarnDRTwoInvSE}{10}

\newcommand{\NoHarnDRTwoInvLI}{7}
\newcommand{\NoHarnAffectedLI}{15}

\newcommand{\NoHarnAffectedStale}{10}

\begin{document}
\title[Structure Before Intelligence]{Structure Before Intelligence: Gating
What Agentic Scientific Workflows Release}

\author{Ashish Acharya}
\email{aacha003@odu.edu}
\affiliation{\department{Department of Computer Science}\institution{Old Dominion University}\city{Norfolk}\state{Virginia}\country{USA}}
\author{Aman Al Amin}
\email{malam007@odu.edu}
\affiliation{\department{Department of Computer Science}\institution{Old Dominion University}\city{Norfolk}\state{Virginia}\country{USA}}
\author{Hong Qin}
\authornote{Corresponding author.}
\email{hqin@odu.edu}
\affiliation{\department{Department of Computer Science}\department{School of Data Science}\institution{Old Dominion University}\city{Norfolk}\state{Virginia}\country{USA}}

\begin{abstract}
An agentic scientific workflow can run to completion and still release
wrong results: a silent numerical error, a stale input, or a defect in a
shared upstream step can contaminate several outputs while a sibling branch
stays correct. We treat trust in such results as a \emph{release decision}.
A harness around the workflow's directed acyclic graph (DAG) decides which
outputs may be used: it enforces mandatory checks, contains failures to the
affected branch, recomputes from tracked versions, adds validators, and
releases a result only if a risk model estimates its probability of being
invalid as low. Language-model agents work inside the harness, choosing
corrective actions and reviewing releases, but cannot override it. On
\ResInstances\ held-out paired instances of two workflows with injected
faults, we switched components on and off. Each supplied a different
property. Containment and recovery raised the number of correct results.
Validation supplied safety: fixed rules without validation released
\ResTwoInv\ invalid results, while validating every candidate released
none. Risk estimation supplied efficiency, matching that safety with
\AblFourVsVValPct\% fewer validators at somewhat lower yield, although
better targeting than random validation was not shown. Lineage evidence
helped on one workflow. The language model added no measurable benefit:
whether invalid results escaped depended on the harness, not the model.
This proof of concept points to a practical path: placing agents inside
explicit release-control structure can keep released scientific results
trustworthy, and it gives more capable agents a measurable baseline to
improve on.
\end{abstract}


\maketitle

\section{Introduction}
\label{sec:intro}
A computational analysis can finish without error and still support the
wrong scientific conclusion. A parameter can be transcribed incorrectly, a
summary can be computed from a superseded input, or a numerical solver can
return a trajectory whose accuracy is insufficient for the quantity
reported. Execution completion, file integrity, and artifact availability are
necessary, but none of them establishes that a released number satisfies
the scientific task. The gap matters more as language-model agents plan and
execute analyses and as published papers are turned into executable
agents~\cite{miao2026paper2agent}: more computation is produced with less
direct human inspection of intermediate results.

Scientific workflows share ancestors. A defect
in a shared normalization or parameter file can contaminate several
deliverables, while a sibling branch remains valid and worth releasing.
Evidence also arrives over time: a late diagnostic, or a correction to a
source after release, can change which artifacts should be trusted. Two simple responses fail. Halting the
whole workflow at the first failed check avoids invalid release but discards
valid work. Validating every artifact exhaustively is expensive and still
leaves the question of what to do when a check fails.

\begin{figure}[t]
\centering
\begin{tikzpicture}[
  font=\footnotesize, >=stealth,
  wn/.style={circle, draw, minimum size=5mm, inner sep=0pt},
  dv/.style={wn, double, double distance=0.6pt},
  bx/.style={draw, rounded corners=1.5pt, align=center, inner sep=2.5pt,
    fill=white},
  ar/.style={->, thick},
  lb/.style={font=\scriptsize, inner sep=1pt},
]
\node[wn, fill=red!25, draw=red!70!black] (S) at (0.3,0.65) {$S$};
\node[wn] (a1) at (1.15,1.25) {};
\node[wn] (a2) at (2.0,1.25) {};
\node[dv] (A) at (2.85,1.25) {$A$};
\node[wn] (b1) at (1.15,0.05) {};
\node[wn] (b2) at (2.0,0.05) {};
\node[dv] (B) at (2.85,0.05) {$B$};
\foreach \u/\v in {S/a1,a1/a2,a2/A,S/b1,b1/b2,b2/B} \draw[->] (\u) -- (\v);
\node[lb, text=red!70!black, anchor=north] at (S.south) {fault};
\node[lb, anchor=south west] at (-0.2,1.55) {workflow DAG $H_t$};

\node[bx, fill=green!8] (man) at (6.55,0.65)
  {\textbf{release manifest}\\released versions;\\retractions recorded};

\draw[dashed, rounded corners=3pt, fill=blue!3]
  (-0.3,-0.5) rectangle (8.15,-2.95);
\node[lb, anchor=south east, text=blue!50!black] at (8.1,-2.9)
  {\textbf{release-control layer}};
\node[bx] (risk) at (1.35,-1.2)
  {risk $p_t(d)$ from\\checks + lineage};
\node[bx] (test) at (4.05,-1.2) {release iff\\$p<p^\ast$};
\node[bx] (act) at (3.55,-2.3) {validate, recompute,\\or withhold};
\node[bx, fill=orange!12] (agent) at (6.6,-1.45)
  {LM agent: picks\\a fix from a menu;\\reviews releases};

\draw[ar] (1.35,-0.25) -- node[lb, right, pos=0.15] {candidate} (risk);
\draw[ar] (risk) -- (test);
\draw[ar] (test.north) |- node[lb, pos=0.75, above] {yes: release} (man.west);
\draw[ar] (test.south) -- node[lb, right] {no} (test.south |- act.north);
\draw[ar] (act.west) -| node[lb, pos=0.3, below] {new version} (-0.1,0.65)
  -- (S.west);
\draw[ar, dotted] (agent.south) |- (act.east);
\draw[ar, dotted] (agent.west) -- (test.east);
\end{tikzpicture}
\caption{The harness as a release gate: it decides which outputs may be
used. A fault in shared ancestor $S$ can
reach both deliverables $A$ and $B$. For each candidate the controller
estimates $p_t(d)$ from check evidence and, optionally, lineage; it
releases iff $p<p^\ast$ and otherwise validates, recomputes, or withholds.
Only the manifest states what may be used, and later evidence is recorded
as a retraction. The language-model agent acts inside the layer and cannot
override the release rule.}
\label{fig:overview}
\end{figure}
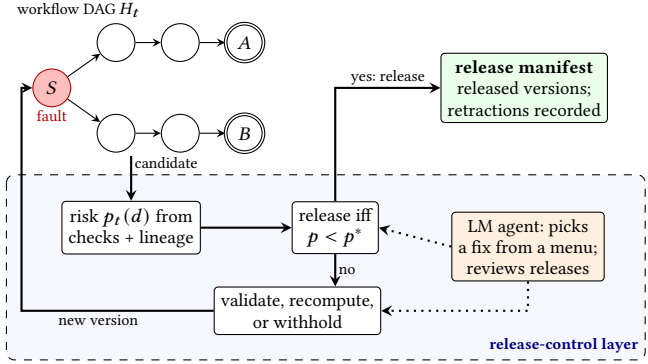

We therefore treat trust in such results as a \emph{release decision} and
study a directed acyclic graph (DAG) harness that makes it. During execution, a controller enforces mandatory checks and release
closure, contains failed branches, and decides which validators to insert,
which artifacts to retry, repair, or recompute from approved inputs, and
which deliverables to release, withhold, or retract. A release manifest, not
the presence of an output file, is the authoritative statement of what a
consumer may use. In the risk-guided policies a fixed model estimates the
probability that the current candidate is invalid, from runtime evidence
and, in one variant, its actual \emph{lineage} (evidence about ancestors
along the versions actually consumed); a candidate is released only if
this risk is below a threshold $p^\ast$ set from contract costs. Language-model
agents operate \emph{inside} the harness: they choose corrective actions and
review releases, but cannot override mandatory checks or the release rule
(Figure~\ref{fig:overview}).

This is a proof of concept. With the same contract, validator library,
action costs, and budget caps (spent cost differs by policy and is reported,
not matched), we ask which harness component supplies which property: do
containment and recovery raise valid yield; does validation prevent invalid
release; does risk estimation reduce the validation needed, and does lineage
help beyond local evidence; and does a language model inside the harness
change any of these? In our experiments the answer was structural:
reliability came from the harness, not from the model. The scope is narrow: two
workflows, \ResCaseGroups\ evaluation case groups, synthetic faults, one
published source study, and disclosed post-hoc addenda
(Section~\ref{sec:threats}). Dependency tracking, conditional execution,
runtime enforcement for agents, and selective prediction all exist
(Section~\ref{sec:related}); the contribution is a controlled decomposition
of how they combine into a release decision when the outcome is scientific
validity, labeled by oracles separate from the runtime checks.

\paragraph{Contributions.} Within the scope stated in
Section~\ref{sec:threats}:
\begin{enumerate}
\item \emph{Release control as a harness.} A formulation that separates
scientific validity from completion, integrity, and availability, defines
deliverable risk $p_t(d)$ conditional on runtime evidence and actual lineage
(Section~\ref{sec:problem}), and releases a candidate iff a fixed risk
estimate satisfies $p<p^\ast$ (Section~\ref{sec:risk}).
\item \emph{Fixed controller semantics} in which mandatory checks, typed
dependencies, persistent issues, bounded graph changes, versioned recovery,
release closure, and retraction hold for every policy, so that policies
differ only in validation and recovery choices
(Section~\ref{sec:controller}).
\item \emph{An agent inside the harness.} A language model proposes
corrective actions from a fixed menu and reviews releases; malformed or
off-menu output becomes abstention, and every call is logged and replayable
(Section~\ref{sec:agentic}).
\item \emph{A paired proof-of-concept benchmark} of two source-linked
workflows, six scenario families, five deterministic policies, agentic
variants on three open-weight models (two paid models in a supplementary
addendum), post-hoc validation ablations, and outcome
oracles separate from the runtime, with predefined denominators and
case-group-disjoint splits (Section~\ref{sec:design}).
\item \emph{A per-workflow decomposition} (Section~\ref{sec:results},
Table~\ref{tab:takeaway}). Containment and recovery raised valid yield;
validation, not risk estimation, removed invalid releases; risk estimation
reduced the validation needed, although better targeting than random
placement was not shown; lineage helped in one workflow through one
mechanism; and a language model inside the harness changed none of these.
\end{enumerate}

\section{Related Work}
\label{sec:related}
\paragraph{Paper-derived agents.}
Paper2Agent converts a research paper and its codebase into a
tool-invoking agent: multiple agents build a Model Context Protocol server
exposing the paper's tools and workflows, generate and run tests to make it
robust, and validate that the resulting agents reproduce the original
papers' results~\cite{miao2026paper2agent}. Building
tools from papers, testing them, and running agentic scientific workflows are
therefore not new here. We assume the task and workflow are already
specified and study a narrower question: how runtime evidence should govern
validation, recovery, and release within one execution. We do not reproduce
Paper2Agent's evaluation or assume its tools lack validation.

\paragraph{Workflow engines and dynamic execution.}
Snakemake~\cite{koster2012snakemake} and Nextflow~\cite{ditommaso2017nextflow}
organize reproducible computation through declared dependencies. Snakemake
checkpoints re-evaluate the job DAG after a checkpoint's outputs are
produced, enabling data-dependent conditional execution~\cite{snakemakecheckpoints},
and rules can declare output-property checks~\cite{snakemakevalidation}.
Dynamic graphs and user-defined validation are thus established features.
Our baselines receive the same validators, and our object of study is how
validation results, lineage, and risk estimates determine release, not
whether an engine can run a check. Provenance models such as
PROV-DM~\cite{moreau2013prov} supply concepts for derivation and
invalidation that our execution history follows; we do not propose a new
provenance model.

\paragraph{Provenance, recomputation, and data validation.}
CWL~\cite{crusoe2022cwl} and Galaxy~\cite{galaxy2024} make workflows
portable and record what ran. noWorkflow captures provenance of
scripts~\cite{murta2015noworkflow}, and ReproZip packages an execution for
re-running~\cite{chirigati2016reprozip}. Build systems such as make and
Bazel recompute only out-of-date targets; Mokhov et
al.\ formalize this family of systems~\cite{mokhov2018build}. Our
invalidate-and-recompute action reuses this idea, but it is triggered by
scientific evidence rather than by changed inputs. TensorFlow Data
Validation~\cite{caveness2020tfdv} and Great
Expectations~\cite{greatexpectations} check data against declared schemas
and expectations. Such checks play the role of our validators. We study
when to run them and how their results gate release.

\paragraph{LLM science agents and benchmarks.}
The AI Scientist automates an ML research loop end to end~\cite{lu2024aiscientist},
and Biomni is a general-purpose biomedical agent~\cite{huang2025biomni}.
BixBench~\cite{mitchener2025bixbench}, CORE-Bench~\cite{siegel2024corebench},
and PaperBench~\cite{starace2025paperbench} measure agents on computational
biology analyses, computational reproduction of published research, and
replication of ML papers. These benchmarks score task outcomes. We hold the
workflow fixed and score which artifacts are released.

\paragraph{Scientific agent control.}
ControlA identifies detection, containment, recovery, and provenance as
requirements for reliable agentic science and connects workflow
instrumentation with agent safeguards~\cite{gueroudji2025controla}. Its
motivation is close to ours. Our study contributes a comparative evaluation
of specific release policies against oracle-labeled scientific
outcomes, with cost reported, rather than an architecture.

\paragraph{Runtime enforcement and information flow.}
Shield synthesis enforces specified properties at runtime while minimally
interfering with a system~\cite{bloem2015shield}. AgentSpec provides a
language of triggers, predicates, and enforcement actions for LLM
agents~\cite{wang2026agentspec}. Sherlock selects where to verify within an
agentic workflow and uses speculative execution with
rollback~\cite{ro2025sherlock}; topology-aware and cost-aware verification
with recovery is therefore prior art. CaMeL separates control from data flow
to defeat prompt injection~\cite{debenedetti2025camel}, and Fides tracks
integrity and confidentiality labels through agent
actions~\cite{costa2025fides}. These systems inform our mandatory blocking
and containment semantics, which every policy shares. Our evaluation target
differs: scientific validity of released deliverables and the useful partial
release of unaffected branches, not security properties. We do not claim
those systems could not be extended to this setting.

\paragraph{LLMs as reviewers.}
Strong LLM judges can agree with human preferences, but they show
position, verbosity, and self-enhancement biases~\cite{zheng2023judge}.
Changing the order of the candidates alone can change an LLM evaluator's
verdict~\cite{wang2024fair}. Our reviewer therefore only gates release; it
cannot release a blocked version. We report its agreement with the oracle
per model.

\paragraph{Selective prediction and risk control.}
Selective classification trades coverage for lower error by
abstaining~\cite{geifman2017selective}. Conformal risk
control~\cite{angelopoulos2024crc} and Learn-then-Test~\cite{angelopoulos2025ltt}
give finite-sample guarantees on expected monotone losses or risk under
exchangeability-type assumptions. Release with withholding is a selective
decision, so these results frame our metrics: invalid releases per released
output is a conditional (selective) risk, while invalid releases per requested
output is unconditional. Our small logistic model, fit to a synthetic fault
mixture with adaptive, correlated, policy-dependent episodes, does not
inherit those guarantees, and we report calibration only as an empirical
property of the benchmark.

\section{Problem Formulation}
\label{sec:problem}
\paragraph{Task contract.}
A task contract fixes, before execution, the scientific target, evidence
tier (arithmetic over a published table, or synthetic parameters), source
identities by content hash, the logical nodes and primary deliverables
$D$, units and tolerances for each deliverable, the mandatory and optional
checks with declared costs, the admissible actions, approved computational
alternatives, and budget caps (attempts per node, total repairs, total cost).
No policy or model may change the target or a tolerance to make a candidate
pass.

\paragraph{Execution history.}
Execution produces an append-only directed acyclic graph $H_t$ of attempts,
artifact versions, check records, and decision events. Each artifact version
$v$ has a unique identity, a content digest, a producing attempt, and the
versions it actually consumed, each tagged as a \emph{scientific input}
(values consumed as accepted inputs) or a \emph{diagnostic control}
(inspected, never consumed). A retry or repair always creates a new version;
no identity is reused. A logical deliverable $d \in D$ may therefore have
several versions; $\mathrm{lin}_t(v)$ denotes the scientific-input ancestors
of $v$ in $H_t$.

\paragraph{Validity versus other properties.}
A version $v$ of deliverable $d$ is \emph{scientifically valid} if its value
satisfies the contract and tolerance for $d$, as judged by a
separate oracle after execution. This is distinct from:
\emph{completion} (the producing attempt terminated successfully),
\emph{integrity} (bytes match the recorded digest and consumed versions are
the recorded ones), and \emph{availability} (a version exists and can be
released). A completed, intact, available artifact can be invalid; a valid
artifact can be withheld. Oracle verdicts are \emph{valid}, \emph{invalid},
or \emph{unobserved} when a candidate is absent or the oracle's applicability
conditions do not hold; unobserved is never imputed.

\paragraph{Deliverable risk.}
For deliverable $d$ at time $t$, let $\mathcal{E}_t$ be the evidence
observable at runtime (check records, diagnostics, issues, decision events)
and $\mathrm{lin}_t(d)$ the actual lineage of the current candidate. We define
\begin{equation}
p_t(d) \;=\; \Pr\bigl(\text{candidate for } d \text{ is scientifically
invalid} \;\big|\; \mathcal{E}_t,\, \mathrm{lin}_t(d)\bigr).
\label{eq:risk}
\end{equation}
The probability is with respect to the benchmark's declared fault mixture,
not real-world failure prevalence. Fault labels, scenario identities, and
oracle outputs are not part of $\mathcal{E}_t$.

\paragraph{Release control.}
A policy maps the observable state to an admissible action: run, insert a
validator, retry, select an approved alternative, repair, invalidate and
recompute descendants, release, retract, abstain, or stop globally. An
episode ends with a release manifest listing released version identities,
and a record of any later retractions. The objective trades three
quantities: invalid releases, valid deliverables withheld, and resource cost.
Section~\ref{sec:design} fixes the corresponding metrics and denominators.

\section{Controller Semantics}
\label{sec:controller}
All policies run on one controller. The rules below are fixed and cannot be
overridden by any policy or risk estimate; policies choose only among the
actions these rules admit.

\paragraph{Versions and states.}
History is append-only. Every attempt (run, retry, alternative, repair,
recompute) creates a fresh version with its own identity and producing
attempt, in one of the states \emph{staged}, \emph{accepted},
\emph{blocked}, \emph{abstained}, \emph{released}, \emph{superseded}, or
\emph{retracted}. Failed versions are kept. Repair is an attempt outcome,
not a state. The history graph is asserted acyclic at the end of every
episode.

\paragraph{Blocking and issues.}
A mandatory check that fails, is missing, has no implementation, or raises
(\emph{unknown}) blocks the version whatever the policy or risk; a
non-candidate ``accept'' is forced to abstain. A consumed input that is a
superseded version raises a version issue from provenance. Issues propagate
without attenuation to all scientific-input descendants under a shared
issue and cause identifier. An issue is cleared only by a resolution record
naming one of two rules, \emph{superseded by a fresh version} (evidence:
the accepted replacement) or \emph{source update applied} (evidence: the
scheduled update), and the evidence must exist in the history.

\paragraph{Containment.}
Blocking propagates along scientific-input edges only. Diagnostic-control
edges (a validator or repair inspecting a version) are recorded for
provenance but never make the inspected version a scientific ancestor.
\DR{0} stops the whole episode at the first blocker; deliverables accepted
before the stop are still released, so \DR{0} depends on branch order, which
each scenario fixes. \DR{1} contains the failure to its branch.

\paragraph{Bounded graph changes.}
Dynamic policies may change the graph only by the contract's admissible
actions: insert a validator (run declared optional checks), retry, select an
approved alternative, repair from approved inputs, or invalidate and
recompute. Each change is checked for acyclicity before it is applied and
recorded with the graph before and after; the graph revision only increases.
Budgets cap attempts per node, total repairs, and total cost. When a cap
prevents a wanted recovery, the controller records an abstention with
outcome \emph{exhausted}.

\paragraph{Release and late evidence.}
For dynamic policies, ``accept'' at a candidate admits a version;
deliverables are released only at closure. Closure rechecks, along the
whole scientific lineage, stored-byte integrity against digests,
current-version consistency, mandatory checks, and unresolved issues, with
at most three recovery rounds; a deliverable that still fails is withheld.
Late evidence is invalidating only and attaches to the target's logical
node: before release it invalidates the affected lineage, which is
recomputed only when the evidence carries a corrected runtime input and is
otherwise withheld; after release it is recorded as a retraction. Retracted
versions stay counted as released. Scheduled events fire at a logical
checkpoint and occurrence that counts first-run productions only, so
recovery attempts never shift them.

\paragraph{Policy boundary and fixed estimators.}
Policies see runtime observations only; no policy receives the scenario
identity, family, or event schedule, and a test checks that decisions do not
change across scenario identities. Risk policies call one estimator
interface (Section~\ref{sec:risk}). If the estimator cannot produce a risk,
the episode ends with status \emph{error}; there is no fallback to another
policy. Before loading, the risk model, its benefit table, and its loss
weights are checked against pinned hashes, and any mismatch stops the run.
Each episode manifest records which model, table, and weights ran. A side
record, hashed into the manifest, keeps the full local and ancestral
feature vector of every consultation under every policy.

\paragraph{Isolation boundary.}
Runtime code receives only the scenario package (inputs, runtime-visible
events, branch order). Fault labels, expected values, and oracle verdicts
are stored separately and joined to runtime records by version identity only
after an episode; no runtime module imports the oracle, which tests enforce
statically and dynamically. This is a research-harness boundary, not an
operating-system sandbox.

\section{Risk Model and Action Selection}
\label{sec:risk}
Two policies use the risk model: \DR{3} sees local evidence about the
candidate only, and \DR{4} also sees its lineage.

\paragraph{Features.}
Features come from one source, the controller's declared feature functions,
so policy and model cannot drift apart. \DR{3} uses 18 local features of the
candidate: a check summary (including the maximum normalized residual
$|\text{error}|/\text{tolerance}$), attempt and recovery state, and
unresolved-issue counts by kind. \DR{4} adds 9 ancestral features over the
actual scientific-input lineage, in which each shared issue cause is
counted once (full list in Supplement~S3). Diagnostic
edges are not traversed, and node reliabilities are never multiplied as if
independent. A missing residual is flagged by its availability indicator
and filled with a fixed $0$, never imputed from labels. A leakage
guard rejects hidden metadata, oracle records, scenario or case identifiers,
and future events. Nine features are constant on all development rows and
carry no information: unknown-check count, infrastructure error, four issue
kinds (execution, integrity, closure, input), and three ancestral fields
(fails, unknowns, distinct causes). \DR{3} thus effectively uses 12 of 18
local features, and \DR{4} 6 of its 9 additional ones.

\paragraph{Training and calibration rows.}
A row is a labeled policy consultation in any situation (candidate, blocked
after a mandatory failure, or invalidated by late evidence); a version
checked before and after a validator gives two rows. Development rows come
from \DR{2} and a development-only exploration variant of it that, with
probability $0.2$, picks a random admissible recovery or, where \DR{2} would
accept, inserts the validator instead. Calibration rows come from \DR{2}
only. The development fit uses 2000 rows (636 invalid) from
33 case groups; calibration uses 477 rows (154 invalid) from 16 disjoint
groups. Unobserved labels are excluded and counted; none occurred.

\paragraph{Estimator and calibration.}
Both models are $\ell_2$-regularized logistic regressions (penalty $1.0$,
L-BFGS-B, standardization fit on development rows only, deterministic
refit). The calibrator is the identity on the raw output. Thresholds
fixed before any fit require at least 20 fitting rows and 30 calibration
rows with at least five per class, and permit the word ``probability'' only
if expected calibration error (ECE) on the calibration split is at most
$0.10$ and the Brier score beats the prevalence reference. Both models pass
on the full calibration set: ECE $0.020$ (\DR{3}) and $0.011$ (\DR{4});
Brier $0.190$ and $0.184$ against a reference of $0.219$. We therefore call
the output a probability, conditional on the benchmark's fault mixture.
On the evaluation split this holds for the risk at a version's final
consultation, which usually includes validator evidence
(Section~\ref{sec:results}); we use the word only in that sense.
Discrimination is modest. On the 459 candidate rows, AUC is $0.62$ (\DR{3})
and $0.65$ (\DR{4}). The overall AUC ($0.67$, $0.69$) is inflated by blocked
and invalidated rows, which are all invalid (10 and 8 rows, one class, below
the support threshold; disclosed, not gating). The fitted models and
their hashes are released with the artifact.

\paragraph{One-step action criterion.}
Let $p$ be the estimated risk of the version under decision, $c(a)$ the
measured action cost, and $\lambda_I$, $\lambda_W$ the costs of an invalid
release and of leaving an eligible deliverable without a valid release. The
loss is
\begin{equation}
\begin{aligned}
L(\text{abstain}) &= \lambda_W, \qquad
L(\text{accept}) = (\lambda_I+\lambda_W)\,p,\\
L(a) &= c(a) + \min\bigl((\lambda_I+\lambda_W)\,p',\ \lambda_W\bigr),\\
p' &= p\,(1-f_a) + (1-p)\,b_a,
\end{aligned}
\label{eq:action}
\end{equation}
for a corrective action $a$ with fix rate $f_a$ and break rate $b_a$. A
validator's loss is $c+\sum_o \Pr(o)\min((\lambda_I+\lambda_W)p_o,
\lambda_W)$ over outcomes $o\in\{\text{flag},\text{pass}\}$, with Bayes
posteriors $p_o$ from its sensitivity and false-alarm rate. The policy takes
the argmax of $L(\text{abstain})-L(a)$ over admissible, supported actions;
abstention scores $0$ and wins ties. At a candidate it compares accept,
validator, and abstain. Accept beats abstain iff
$p<p^\ast=\lambda_W/(\lambda_I+\lambda_W)$. Weights are anchored to the
contract rather than to behavior: $\lambda_W$ is the contract cost
budget per deliverable ($60/2=30$ for both workflows) and
$\lambda_I=5\lambda_W=150$, so $p^\ast=1/6$; terminal costs are $0$.
Mandatory blockers apply before Eq.~\eqref{eq:action}, and no loss can
release a blocked version. The loss charges intermediate candidates with
the deliverable-level $\lambda$, which only approximates downstream value.

\paragraph{Action benefits.}
Risk alone does not identify which repair works. Fix, break, sensitivity,
and false-alarm rates are estimated per action and context from development
action records (Supplement~S3). One support rule, set before the fit,
covers every action: an action without enough labeled records is not
scored, and the policy abstains rather than take it. Where valid-before
records are scarce, a worst-case bound replaces the break or false-alarm
rate; because the loss is non-decreasing in it, the bound can only lower the
benefit of recovery, biasing against \DR{3} and \DR{4}. Table and weights
are identical for the two risk policies, which differ only in features.

\section{Agentic Proposer and Reviewer}
\label{sec:agentic}
The agent operates inside the release-control layer. A language model is
placed in two bounded roles on top of a base policy (\DR{0}, \DR{2}, or
\DR{4}), giving the arms \Ag{0}, \Ag{2}, and \Ag{4}. The base policy
still chooses the action class (validate, correct, release, abstain, or
stop), and every controller rule of Section~\ref{sec:controller}, including
the risk release rule under \DR{4}, still applies. The question is whether
the agent changes valid yield or invalid release relative to the same base
policy, with cost reported alongside.

\paragraph{Proposer.} When the base policy chooses to correct, the model
picks the concrete corrective action and its declared parameters from a
per-workflow \emph{proposal menu}.
Each option is a recovery action (retry, select an approved alternative,
repair, or invalidate and recompute) applied to the node under decision.
The menu is intersected with the actions admissible at that point. Solver
settings are reachable only through the contract's approved alternative
solvers, and declared parameters are recorded but not executed (menus in
Supplement~S2). Admissibility, budgets, and acyclicity are checked exactly as
for any other action.

\paragraph{Reviewer.} By construction, the reviewer cannot reject a
candidate it sees. The controller consults it only for a candidate with no
unresolved issue (a failed check or a mandatory unknown has already blocked
release), and for exactly this state the prompt instructs approval. The
model returns a verdict (approve, reject, or uncertain, with a reason),
recorded as a review check; release requires \emph{approve}, and any other
verdict blocks release in that round, after which the base policy may
correct, validate, or abstain. A reject path exists in code and the prompt
permits rejection on contradictory evidence, but the reviewer approved
every candidate it saw (Section~\ref{sec:results}). It can only withhold,
never release a version the controller blocks, and its verdict is not a
risk-model feature. We make no safety claim for it; its approval rate
reflects this structure, not reviewer quality, so the agentic arms test the
proposer. Because \DR{0} never invokes the proposer, \Ag{0} is \DR{0} plus
a reviewer gate.

\paragraph{Evidence and output.} Prompts contain only runtime-visible
evidence: contract text, check results, open issues, a lineage summary, and
remaining budgets. The leakage guard scans every prompt for scenario
identifiers, family names, fault labels, and oracle fields. A test checks
that decisions do not change when scenario identifiers are renamed.
Responses are JSON, constrained by a schema where the server supports it.
Output that cannot be parsed, an off-menu choice, a timeout, or a server
error is recorded as an error and becomes abstention for that decision. A
reviewer error also blocks release. Errors are never repaired, and there is
no deterministic fallback.

\paragraph{Models, audit, and cost.} The panel has three open-weight models,
Llama~3.3 70B, Gemma~3 27B, and Mistral Small 3.2 24B, served locally by
Ollama with \texttt{\LlmQuant} weights and pinned by manifest digest.
Decoding uses temperature \LlmTemperature, a fixed seed, a context of
\LlmNumCtx\ tokens, at most \LlmNumPredict\ output tokens, and no retries
(Supplement~S2). Every call is appended to a per-episode call log (role,
prompt, response, model digest, options, timing, tokens) whose hash enters
the episode manifest. A replay adapter re-executes an episode from its log
without a GPU, and an automated audit by a separate read-only agent replayed
every agentic episode byte-identically. Re-running a model is not a
substitute: in a check outside the harness (16 development prompts, six
calls each), decisions were stable on repeated calls but output text was
not byte-reproducible, because Ollama's prompt-cache state depends on call
order. Prompts are written from contract and action semantics only,
piloted on development, sanity-checked on calibration, and never tuned on
evaluation. One revision of the reviewer prompt was allowed, on development
data only: under the first version the reviewer answered ``uncertain''
whenever optional validators had not run, so \Ag{2}, which inserts none,
made no releases in the development pilot. The revision gives the reviewer
the contract's required and optional checks. A missing optional check
alone is no longer grounds for ``uncertain'' or ``reject''; ``uncertain'' is
reserved for missing or failed required checks and contradictory evidence.
The revision has no workflow-, fault-, or scenario-specific wording, and
both versions are kept (Supplement~S2). Actions are charged to contract
cost as under every policy, and a reviewer call is not charged as a
validator; calls and tokens are reported separately and counted from the
call logs (the runner's own summaries omit tokens; Supplement~S6). A later
addendum with two paid API models is reported only in Supplement~S7.

\section{Experimental Design}
\label{sec:design}
\subsection{Workflows and Oracles}
Each workflow has seven logical nodes: one shared upstream node and two
three-node branches, each ending in a deliverable (Table~\ref{tab:cases}).
Contracts, targets, and tolerances were fixed by content hash before any
scenario was generated; no event or alternative changes them.
\emph{Published table.} From a hash-verified transcription of Table~1 of a
published evaluation of variant-calling pipelines~\cite{zanti2021} (28
conditions, five strata), branch A reports mean Matthews correlation per
calling padding over the nine non-commercial conditions; branch B ranks all
28 conditions. A runtime query selects the strata. All instances derive from
this one source study.
\emph{SIR model.} Two branches with different transmission rates integrate
the closed, frequency-dependent Kermack--McKendrick
model~\cite{kermack1927} from a verified registry of synthetic rates and
initial fractions; each summary reports $R_0$, peak size and time, final
size, and attack fraction.

\begin{table}[t]
\caption{Workflow contracts (both: 7 logical nodes, at most 3
attempts per node, 6 repairs, and cost 60).}
\label{tab:cases}
\small
\begin{tabularx}{\linewidth}{@{}lYY@{}}
\toprule
 & Published table & SIR model \\
\midrule
Evidence tier & Published values & Synthetic parameters \\
Shared node & source table & model specification \\
Deliverables & A report, B report & summary A, summary B \\
Checks (mand./opt.) & 3 / 5 & 4 / 4 \\
Alternatives & rebuild from verified source; decimal mean & load from registry; 3 tighter solvers \\
\bottomrule
\end{tabularx}
\end{table}

\emph{Oracles.} The oracles label every version of every logical
node, intermediates included; runtime working copies are never read. Table
references are exact decimals recomputed from the oracle's own hash-verified
copy of the source table. SIR references are analytical (phase-plane
invariant, final-size equation, peak, and a quadrature time map) and never
integrate the ODE; full definitions are in Supplement~S4. A field outside
its applicability conditions is \emph{unobserved}. A version is invalid if any field fails or
is malformed, unobserved if any remaining field is unobservable, and valid
otherwise. Intermediate fields borrow the tolerance of the deliverable field
they feed, since contracts set none. A label states whether a version's
\emph{checkable content} satisfies the contract. A defect in a
setting the contract leaves unconstrained (here a loose SIR solver
tolerance) is thus labeled valid where it originates and invalid where it
appears downstream; this affects 9 intermediate versions in development and
calibration. Labels never come from generator metadata. On development and
calibration, oracle verdicts agree with the generator's deviation labels on
all 600 checked deliverable versions, and no node has an unobserved label.

\subsection{Scenario Families}
Each workflow has six families, each instance accepted only after a static
reference run against the clean baseline of its case group: \emph{clean} (no
events, alarms, or deviations); \emph{benign} (a semantically equivalent
input variant, such as reformatting, units, or a more accurate solver, with
no alarms or deviations); \emph{silent error} (exactly one deliverable
deviates, no mandatory alarm, at least one optional alarm);
\emph{stale} (a superseded upstream version is consumed, which the
controller detects from provenance rather than from checks);
\emph{shared ancestor} (both deliverables deviate through a shared node or
input, no mandatory alarm); and \emph{late invalidation} (an invalidating
notice arrives during the run or after release; 6 of 10 in-run in
evaluation). Late evidence is never confirming. For every faulted instance an
approved recovery restores all deliverables within tolerance. Events attach
to logical checkpoints, so added attempts do not shift them across policies.

\subsection{Policies}
All policies share contracts, mandatory checks, validator library, costs,
and budgets (Table~\ref{tab:policies}). \DR{0} and \DR{1} run every declared
check, optional ones included, as strong-check static baselines; the cost of
checking everything is reported. The dynamic policies (\DR{2}, \DR{3}, \DR{4}) run mandatory checks and add
optional ones only by inserting a validator; they share admissible actions
and caps. \DR{2} inserts all available validators when any normalized
residual reaches $0.5$, and recovers a blocked version by a fixed ladder:
retry an infrastructure error; recompute after a version, integrity, late,
or closure issue; else select an untried approved alternative; else repair;
else abstain. Its validator trigger never fired on development (0
insertions in 120 \DR{2} episodes) or calibration (0 of 477 consultations in
60 episodes), because mandatory-check residuals stayed below threshold in
the families as accepted. We report this as a property of the fixed rule;
the threshold was not retuned. Two sensitivity arms rerun both risk
policies with their original loss weights (Supplement~S6). Two post-hoc
ablation arms reuse the rules and recovery ladder of \DR{2} and vary only
validation (Table~\ref{tab:policies}).

\begin{table}[t]
\caption{Policies. $^{\ast}$Post-hoc ablation arms (Section~\ref{sec:protocol}).}
\label{tab:policies}
\small
\begin{tabularx}{\linewidth}{@{}lYY@{}}
\toprule
Policy & Behavior & Comparison enabled \\
\midrule
\DR{0} & Static graph, all checks, global stop & Global intervention baseline \\
\DR{1} & Static graph, all checks, branch containment & Value of containment \\
\DR{2} & Dynamic, fixed-rule validation and recovery & Value of bounded adaptation \\
\DR{3} & Dynamic, risk-guided, local evidence & Value of estimated risk \\
\DR{4} & Dynamic, risk-guided, local + ancestral evidence & Value of lineage \\
\midrule
\ValAll$^{\ast}$ & \DR{2}, validating every candidate & Value of validation \\
\RandMatched$^{\ast}$ & \DR{2}, random validation at \DR{4}'s volume & Value of risk targeting \\
\bottomrule
\end{tabularx}
\end{table}

\subsection{Splits and Sample Size}
The benchmark has 240 instances: 120 development, 60 calibration, and 60
evaluation (10, 5, and 5 per workflow and family). Case groups are disjoint
across splits: the 25 query pairs of the published table split 13/6/6 and
the 40 sampled SIR registries 20/10/10. The evaluation set has $2$
workflows $\times$ $6$ families $\times$ $5$ instances $\times$ $5$ policies
$=300$ primary episodes, i.e.\ $60$ paired instances; the predeclared
reduction to four instances per family was not needed. The two sensitivity
arms add 120 episodes. Repeating a deterministic run does not add an
independent unit. Primary and sensitivity arms make no language-model calls.

\subsection{Agentic Arms}
\label{sec:agentic-design}
The agentic set has \Ag{0}, \Ag{2}, and \Ag{4}, each run with the three
models: \LlmArms\ arms on the same \ResInstances\ evaluation instances
and seeds (\LlmEpisodes\ episodes), paired with the deterministic
episodes by instance and seed. The pre-declared comparisons are each
agentic arm versus its base policy per model, and \Ag{4} versus \Ag{2}
per model. Across models we report the error rate (off-menu output counts
as an error), the reviewer approval rate, reviewer agreement with the
oracle (computed after evaluation), calls, and tokens (Supplement~S1); wall
time is recorded but was not audited and is not reported. A reduction rule
fixed in advance would have dropped arms if the development pilot had
projected too much compute; the projection kept the full set.

\subsection{Protocol and Disclosure}
\label{sec:protocol}
Contracts, scenario instances, splits, policies, risk models, loss weights,
metrics, and primary comparisons were fixed and hashed before any
evaluation episode ran, and nothing was tuned on the evaluation split. The
agentic arms were added after the deterministic episodes had run but before
any evaluation oracle output was computed, and were fixed in the same way
before their own run. Four analyses were decided after evaluation results
had been seen and are \emph{post hoc}: the validation ablation (\ValAll,
\RandMatched), the no-release-control reference, a recomputation of
calibration that corrects a defect in the original analysis, and an
addendum with two paid API models. Tables mark them with $^{\ast}$; they are
never pooled with the planned arms, and no conclusion rests on the paid-API
addendum (Supplement~S7). Changes adopted before evaluation but after
development or calibration behavior had been seen, such as the loss
weights, are listed with the four post-hoc analyses in the deviation log
(Supplement~S6). For every comparison a null result was an acceptable
outcome.

\subsection{Metrics}
Metrics are computed from raw episode manifests and oracle records;
every ratio is reported with numerator and denominator. The primary
eligibility rule counts every contract deliverable: $60\times2=120$
deliverables per policy (60 per workflow). A secondary rule counts only
deliverables outside the structural lineage of the injected fault, 64 per
policy (clean 20, benign 20, silent error 10, stale 10, late 4, shared 0),
to measure collateral loss such as over-containment.
\begin{itemize}
\item \textbf{Invalid per released}: invalid released versions / oracle-labeled
released versions. Undefined, not zero, when nothing is released; released
but unlabeled versions are reported separately.
\item \textbf{Invalid per requested}: invalid released versions / requested
outputs.
\item \textbf{Valid yield}: eligible deliverables with at least one released,
non-retracted, oracle-valid version / eligible deliverables. Eligibility is
identical for all policies, so a policy gains nothing by failing to produce
a candidate.
\item \textbf{False blocking} (clean and benign, 40 deliverables): eligible
deliverables with no released, non-retracted version / eligible
deliverables. A withheld candidate is not counted as incorrect.
\item \textbf{Recovery success}: deliverables targeted by a recovery action
whose final released, non-retracted version is oracle-valid / deliverables
targeted; also attempts, checks, and descendants recomputed.
\item \textbf{Cost}: declared action cost and wall time.
\item \textbf{Calibration} (\DR{3}, \DR{4}): Brier score and ten-bin reliability
with bin counts, for all labeled versions and for the released subset, to
expose policy-induced selection.
\end{itemize}
Retracted versions count as released but never toward valid yield.
Infrastructure failures and episodes ending in \emph{error} are reported
separately, never dropped. Paired differences between policies use
per-instance counts; 95\% percentile intervals resample whole case groups
(ablation and reference intervals resample scenarios) and are
conditional on the two workflows and the fault generator. The primary comparisons, fixed before
evaluation, are \DR{4} versus \DR{3} (value of lineage) and each risk
policy versus \DR{2} (value of risk guidance over fixed rules); \DR{1} versus
\DR{0} and \DR{2} versus \DR{1} are secondary.

\section{Results}
\label{sec:results}
Every number in this section is generated by script from the evaluation
outputs and an automated audit by a separate read-only agent. A difference
$\Delta$ is a mean per paired instance (two requested deliverables) over
\ResInstances\ instances, with a case-group bootstrap 95\% interval over
\ResCaseGroups\ case groups (\ResCaseGroupsPT\ published table,
\ResCaseGroupsSIR\ SIR); ablation and reference intervals resample
scenarios. Intervals are not corrected for multiplicity. All \EpisodesAll\
planned episodes ran to a recorded final status with \InfraErrorsAll\
infrastructure errors, and every released version has an oracle label
(Supplement~S1). Table~\ref{tab:takeaway} summarizes what each component
supplied, Table~\ref{tab:results} and Figure~\ref{fig:results} give the
outcomes, and per-family tables are in Supplement~S1.

\begin{table}[t]
\caption{What each harness component supplied (evaluation split; evidence
in the paragraphs below).}
\label{tab:takeaway}
\footnotesize
\setlength{\tabcolsep}{3pt}
\begin{tabularx}{\linewidth}{@{}>{\raggedright}p{1.35cm}>{\raggedright}p{1.15cm}YY@{}}
\toprule
Component & Supplies & Key evidence & Caveat \\
\midrule
Containment, recovery & valid yield & \DR{1} above \DR{0}; \DR{2} above \DR{1} & the \DR{2} gain came with invalid releases \\
Validation & safety & \DR{2}: \ResTwoInv\ invalid; \ValAll: \AblTwoVInv & highest cost \\
Risk estimation & efficiency & \DR{4}: \ResFourInv\ invalid, \AblFourVsVValPct\% fewer validators than \ValAll & lower yield; targeting not shown \\
Lineage & yield, one workflow & \DR{4} above \DR{3} on the published table only & one mechanism \\
Language model & nothing shown & no agent arm beat its base policy & reviewer cannot reject \\
\bottomrule
\end{tabularx}
\end{table}

\begin{table}[t]
\caption{Deterministic policies, evaluation split. Valid yield is of 60 requested deliverables per workflow; invalid releases and false blocking (clean and benign deliverables not released, of 40) pool both workflows; validators are inserted validator runs (total); cost is mean contract cost per instance. Released versions include those retracted after late evidence (\DR{2} 7 (all invalid), \DR{3} 4, \DR{4} 6, \ValAll{} 7, \RandMatched{} 7; none otherwise), which never count toward valid yield. Every retracted valid version was built on a replacement, made by recovery before release, of the node that a post-release notice named; because a notice attaches to the logical node, not to a version, these retractions were unnecessary and cost valid yield. $^{\ast}$Post hoc (Section~\ref{sec:protocol}).}
\label{tab:results}
\footnotesize
\setlength{\tabcolsep}{3.2pt}
\begin{tabular}{@{}lcccccc@{}}
\toprule
& \multicolumn{2}{c}{Valid yield} & Invalid/ & False & Vali- & \\
\cmidrule(lr){2-3}
Policy & Pub.\ table & SIR & released & block & dators & Cost \\
\midrule
\DR{0} & 26/60 & 22/60 & 0/48 & 0 & 0 & 13.00 \\
\DR{1} & 32/60 & 32/60 & 0/64 & 0 & 0 & 15.50 \\
\DR{2} & 42/60 & 41/60 & 37/120 & 0 & 0 & 11.07 \\
\DR{3} & 2/60 & 49/60 & 0/55 & 20 & 235 & 15.80 \\
\DR{4} & 51/60 & 49/60 & 0/106 & 0 & 248 & 19.65 \\
\midrule
\ValAll$^{\ast}$ & 56/60 & 56/60 & 0/119 & 0 & 359 & 25.17 \\
\RandMatched$^{\ast}$ & 52/60 & 49/60 & 3/111 & 0 & 251 & 21.76 \\
\bottomrule
\end{tabular}
\end{table}

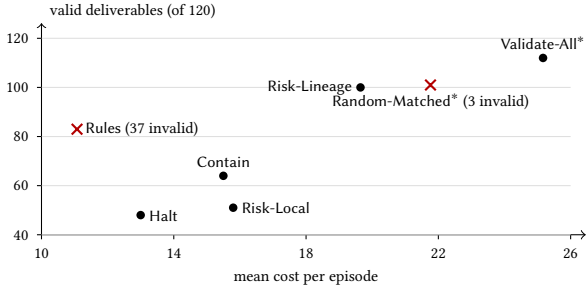
\begin{figure}[t]
\centering
\begin{tikzpicture}[font=\scriptsize]
\draw[->] (0,0) -- (7.2,0);
\draw[->] (0,0) -- (0,2.8000000000000003);
\node[below] at (3.5,-0.35) {mean cost per episode};
\node[above right] at (0,2.75) {valid deliverables (of 120)};
\draw (0.0,0) -- (0.0,-0.06) node[below] {10};
\draw (1.75,0) -- (1.75,-0.06) node[below] {14};
\draw (3.5,0) -- (3.5,-0.06) node[below] {18};
\draw (5.25,0) -- (5.25,-0.06) node[below] {22};
\draw (7.0,0) -- (7.0,-0.06) node[below] {26};
\draw[gray!25] (0,0.0) -- (7.0,0.0);
\draw (0,0.0) -- (-0.06,0.0) node[left] {40};
\draw[gray!25] (0,0.65) -- (7.0,0.65);
\draw (0,0.65) -- (-0.06,0.65) node[left] {60};
\draw[gray!25] (0,1.3) -- (7.0,1.3);
\draw (0,1.3) -- (-0.06,1.3) node[left] {80};
\draw[gray!25] (0,1.95) -- (7.0,1.95);
\draw (0,1.95) -- (-0.06,1.95) node[left] {100};
\draw[gray!25] (0,2.6) -- (7.0,2.6);
\draw (0,2.6) -- (-0.06,2.6) node[left] {120};
\fill (1.312,0.26) circle (1.6pt);
\node[anchor=west, inner sep=3pt] at (1.312,0.26) {\DR{0}};
\fill (2.406,0.78) circle (1.6pt);
\node[anchor=south, inner sep=3pt] at (2.406,0.78) {\DR{1}};
\draw[red!70!black, thick] (0.47-0.07,1.397-0.07) -- (0.47+0.07,1.397+0.07) (0.47-0.07,1.397+0.07) -- (0.47+0.07,1.397-0.07);
\node[anchor=west, inner sep=3pt] at (0.47,1.397) {\DR{2} (37 invalid)};
\fill (2.536,0.358) circle (1.6pt);
\node[anchor=west, inner sep=3pt] at (2.536,0.358) {\DR{3}};
\fill (4.22,1.95) circle (1.6pt);
\node[anchor=east, inner sep=3pt] at (4.22,1.95) {\DR{4}};
\draw[red!70!black, thick] (5.146-0.07,1.982-0.07) -- (5.146+0.07,1.982+0.07) (5.146-0.07,1.982+0.07) -- (5.146+0.07,1.982-0.07);
\node[anchor=north, inner sep=3pt] at (5.146,1.982) {\RandMatched$^{\ast}$ (3 invalid)};
\fill (6.635,2.34) circle (1.6pt);
\node[anchor=south, inner sep=3pt] at (6.635,2.34) {\ValAll$^{\ast}$};
\end{tikzpicture}
\caption{Valid deliverables (of 120) against mean cost per episode for each
arm of Table~\ref{tab:results}. Crosses mark arms that released invalid
deliverables (count in parentheses); $^{\ast}$post hoc.}
\label{fig:results}
\Description{Scatter plot of valid yield against cost for seven arms.}
\end{figure}

\paragraph{Without release control.} A descriptive reference
(post hoc, Supplement~S6) releases every deliverable's first version
unchecked. Because \DR{0} stops at its first mandatory blocker,
\NoHarnNotProdNum\ of \NoHarnNotProdDen\ deliverables were never produced,
so the invalid fraction lies between \NoHarnLowNum/\NoHarnLowDen\
(\NoHarnLow, \NoHarnLowCI) and \NoHarnUpNum/\NoHarnUpDen\ (\NoHarnUp,
\NoHarnUpCI); these are bounds, not an estimate. The generator's labels
mark \NoHarnPointNum/\NoHarnPointDen\ deliverables as fault-affected
(\NoHarnPoint, \NoHarnPointCI), a property of this benchmark's fault mix,
not a natural error rate. Without release control all \NoHarnPointNum\
would be released. This is a reference, not a policy arm.

\paragraph{Containment and recovery supply yield.} \DR{1} released more
valid deliverables than \DR{0} ($\Delta=\DZeroOneValid$ \DZeroOneValidCI),
and neither released an invalid one. \DR{2} raised valid yield over \DR{1}
($\Delta=\DOneTwoValid$ \DOneTwoValidCI) at the price of invalid releases
($\Delta=\DOneTwoInv$ \DOneTwoInvCI). Recovery succeeded in
\ResTwoRecNum/\ResTwoRecDen, \ResThreeRecNum/\ResThreeRecDen, and
\ResFourRecNum/\ResFourRecDen\ attempted recoveries for \DR{2}, \DR{3}, and
\DR{4}.

\paragraph{Validation supplies safety.} The \DR{2} policy's validator
trigger never fired, and it released \ResTwoInv\ invalid deliverables (of \ResTwoRel\
released), all among the \NoHarnPointNum\ fault-affected ones: all
\NoHarnDRTwoInvSA\ shared-ancestor and \NoHarnDRTwoInvSE\ silent-error
deliverables and \NoHarnDRTwoInvLI\ of \NoHarnAffectedLI\ late-invalidation
ones; its \NoHarnAffectedStale\ stale-input deliverables were released
valid. Release closure alone therefore did not stop content faults.
Validating every candidate (\ValAll) removed all invalid releases
(\AblTwoVInv/\AblTwoVRel) and gave the highest valid yield
(\AblTwoVValid/\AblTwoVReq), at the highest cost.

\paragraph{Risk-guided release supplies efficiency; targeting is not shown.}
Against \DR{2}, \DR{4} raised valid yield ($\Delta=\DTwoFourValid$
\DTwoFourValidCI), removed invalid releases ($\Delta=\DTwoFourInv$
\DTwoFourInvCI), and cost \DTwoFourCost\ \DTwoFourCostCI\ more units per
instance; \DR{3} also removed invalid releases ($\Delta=\DTwoThreeInv$) but
lowered valid yield ($\Delta=\DTwoThreeValid$ \DTwoThreeValidCI). The
ablation shows that validation, not risk estimation, removed them. Against
\ValAll, \DR{4} inserted \AblFourVsVValPct\% fewer validators and cost
\AblFourVsVCostPct\% less ($\Delta=\DAblFourVCost$ \DAblFourVCostCI\ units),
at a lower valid-yield fraction ($\Delta=\DAblFourVYield$
\DAblFourVYieldCI). At matched volume (\AblTwoMTwoValidators\ insertions
against \DR{4}'s \AblFourInsertions), random validation released
\AblTwoMTwoInv\ invalid deliverables against \DR{4}'s \ResFourInv, a
difference our \ResInstances\ instances cannot resolve (invalid per released
$\Delta=\DAblFourMTwoInvRate$ \DAblFourMTwoInvRateCI; the interval reaches
$0$ only because resamples that omit the few affected scenarios contain no
invalid release). Risk targeting is therefore not shown; a larger instance
set is the test. \DR{4} was cheaper ($\Delta=\DAblFourMTwoCost$
\DAblFourMTwoCostCI\ units) at similar yield ($\Delta=\DAblFourMTwoYield$
\DAblFourMTwoYieldCI). All \RiskGatedArmsOpen\ risk-gated arms (\DR{3},
\DR{4}, and \Ag{4} with each open-weight model) released
\RiskGatedInvalidOpen\ invalid deliverables; the paid-API addendum is
consistent (Supplement~S7).

\paragraph{Calibration$^{\ast}$.} Scored at the last consultation of each
labeled version (a post-hoc recomputation, Supplement~S6), the risk usually
follows validator evidence, so it describes validator-informed risk, not a
pre-validation prediction: Brier \CalThreeBrier\ \CalThreeBrierCI\ for
\DR{3} (reference \CalThreeRef) and \CalFourBrier\ \CalFourBrierCI\ for
\DR{4} (reference \CalFourRef).

\paragraph{Lineage helps on one workflow.} \DR{4} released more valid
deliverables than \DR{3}, $\Delta=\DThreeFourValid$ \DThreeFourValidCI,
and neither released an invalid one. The whole gain is on the
published-table workflow ($\Delta=\DThreeFourValidPT$
\DThreeFourValidPTCI); on SIR the two tie ($\Delta=\DThreeFourValidSIR$).
On the published table \DR{3} released \ResThreePTValid\ of \ResWfReq\
deliverables valid and withheld all \ResThreePTFB\ clean and benign ones: a
local-only risk model bought safety by over-withholding. The lineage effect
therefore rests on one mechanism in one workflow.

\paragraph{No agent benefit is shown.}
The \LlmEpisodes\ agentic episodes made \AudLlmCalls\ model calls with
\AudLlmErrors\ errors (calls and tokens per model in Supplement~S1). With
Llama and Mistral, every agentic arm matched its base policy exactly on all
\ResInstances\ instances (every paired difference $0$). \Ag{2} released
exactly what \DR{2} released for every model; Gemma's trajectories differed
in \GemmaDRTwoTrajDiff\ scenarios (a repair instead of
invalidate-and-recompute, with the same outputs and cost). The only
non-null difference is Gemma under \Ag{4}, which lowered valid yield
relative to \DR{4} ($\Delta=\DGemmaValid$ \DGemmaValidCI) without releasing
an invalid deliverable; \Ag{4} versus \Ag{2} was then
$\Delta=\DGemmaTwoFourValid$ \DGemmaTwoFourValidCI\ for Gemma, and
identical to \DR{4} versus \DR{2} for the other models. Thus
\AgentModelsOpenIdentical\ of \AgentModelsOpen\ open-weight models
reproduced every base-policy decision; the two paid models of the post-hoc
addendum did as well (Supplement~S7).

The reviewer, which by construction cannot reject what it sees
(Section~\ref{sec:agentic}), approved all \RevN\ candidates, so this arm
tests the proposer. Each model approved \RevDRTwoApprovedInvalid\
oracle-invalid candidates, all under \Ag{2}:
\RevDRTwoInvalidApproved\ were released (the invalid releases of \DR{2})
and \RevDRTwoSuperseded\ were superseded before release when late evidence
arrived during the run. Agreement with the oracle, \RevAgree\ \RevAgreeCI,
equals the share of valid candidates (Cohen's $\kappa=\RevKappa$). With
every model, invalid releases followed the control layer (\ResTwoInv\
under \Ag{2}, none under \Ag{4}), not the model: the language model alone
did not supply reliability.

\section{Limitations}
\label{sec:limitations}
\subsection{Threats to Validity}
\label{sec:threats}
This is a proof of concept, and we make no general claim. We state every
conclusion per workflow; neither workflow is taken to represent scientific
workflows in general.
\begin{itemize}
\item \textbf{Synthetic fault mixture.} Faults, benign changes, and their
frequencies are designed. Risk probabilities, calibration, and every
interval are conditional on this mixture. They are not estimates of
real-world failure prevalence or a distribution-free safety guarantee.
Stale inputs are detected from provenance, and late evidence is never
confirming.
\item \textbf{Small, narrow sample.} There are five evaluation instances per
family and workflow (60 paired instances) from 16 case groups (6 published
table, 10 SIR), and comparisons are not corrected for multiplicity. The
published-table workflow derives from a single study, and the SIR oracle
covers only the closed model.
\item \textbf{Deterministic runs.} Repeating a run adds no independent unit,
so intervals reflect only case-group (or scenario) variation. Agentic arms use one
temperature-0 sample per decision from Ollama's default 4-bit quantized
models. They therefore capture neither sampling variability nor
full-precision behavior.
\item \textbf{Weak fixed-rule baseline.} The \DR{2} policy's validator trigger never
fired, before or on evaluation (\ResTwoValidators\ validators inserted), so \DR{2} acts as a dynamic baseline without
validators. The validation ablation (Section~\ref{sec:results}) indicates that
validation, not risk estimation, removed invalid releases.
\item \textbf{Declared validators.} The same team designed the faults and
the validators, and validators raised no false alarm on development. Safety
assumes the declared validators detect the faults they target, and placement
matters: random validation at matched volume released \AblTwoMTwoInv\ invalid
deliverables.
\item \textbf{Changes after observation.} The loss accounting and weights
and the fit on all situations were adopted after development and
calibration behavior had been observed, though before evaluation
(Supplement~S6); the original weights are run as a sensitivity arm. The
reviewer prompt received one revision on development data, and four
analyses are post hoc (Section~\ref{sec:protocol}).
\item \textbf{Modest risk model.} Candidate AUC is $0.62$ (\DR{3}) and $0.65$
(\DR{4}); higher overall AUC comes from blocked rows, which are almost all
invalid. At candidates the validator dominates: with no false alarm on
development, a pass sharply lowers the posterior. Nine features are constant
on development, so \DR{4} effectively adds 6 ancestral features, not 9. The
one-step loss and worst-case valid-before bound bias against recovery.
\item \textbf{Single mechanism for lineage.} In development smoke runs, \DR{3}
released 0 of 8 published-table deliverables and \DR{4} released 6. At
nodes where no validator is admissible, \DR{3}'s risk exceeds $p^\ast$ and
it abstains. On that workflow, \DR{4}$-$\DR{3} may therefore largely
measure one mechanism.
\item \textbf{Coarse retraction.} A post-release notice attaches to a
logical node, so it also retracts releases that were already built on a
replacement of that node and were valid. This cost the risk and
validation-ablation arms some valid yield (Table~\ref{tab:results}); version-level
attachment would avoid it.
\item \textbf{Labels and audit.} A label states whether checkable content
satisfies the contract. A defect in an unconstrained setting counts
as valid where it originates (9 intermediate versions) and invalid
downstream, and intermediates borrow deliverable tolerances. The oracles
share the task specification and source transcription with production, so
an error common to both would go undetected. Their agreement with the
generator's labels (\LabelAgree/\LabelTotal\ deliverables, development and
calibration only; none was computed on evaluation) is a consistency check,
not independent validation.
\item \textbf{Model involvement.} Language models act only in the agentic
arms, as a menu-bounded proposer and a release reviewer; the arms are not
evidence about open-ended agent autonomy. All \LlmArms\ agentic arms ran
with all three models, with no reduction or fallback. The reviewer is
reached only where its prompt instructs approval, so the arms test the
proposer, not reviewer judgment. The paid-API models (Supplement~S7) are
pinned by endpoint name only, so that addendum is reproducible only by log
replay.
\item \textbf{Tool defects.} Defects found after the runs were left
unfixed in the evaluated code (Supplement~S6).
\end{itemize}

\section{Conclusion}
\label{sec:conclusion}
This proof of concept asked which part of a DAG release-control harness
makes agentic workflow results safe to release. On two workflows with
synthetic faults, the answer was structural. Containment and recovery
supplied yield, validation supplied safety, and risk estimation supplied
efficiency: it kept invalid releases at zero with less validation than
validating everything, but better targeting than random validation was not
shown. Lineage helped on one workflow only, and no benefit from the
language model was shown: invalid releases followed the harness, not the
model. The
ablations are post hoc, and zero observed invalid releases is a count, not
a guarantee.

\ifdefined\arxivbuild
  \begin{acks}
This work was supported in part by an a2 Pilot Award from the University of
Pennsylvania Penn Artificial Intelligence and Technology Collaboratory for
Healthy Aging (PennAITech), an initiative funded by the National Institute
on Aging (NIA) of the National Institutes of Health (Grant No.\
P30AG073105), and by Old Dominion University through MonarchSphere: AI
Incubator, powered by Google Cloud. This work was also supported in part by a grant of access to OpenAI
models through the ChatGPT for Academic Researchers program.
\end{acks}

\else
  \input{sections/availability}
  \input{sections/genai}
\fi

\bibliographystyle{ACM-Reference-Format}
\vfuzz=2pt
{\interlinepenalty=10000 %
\bibliography{references}}

\end{document}


\title{Supplement to: Structure Before Intelligence: Gating What Agentic
Scientific Workflows Release}
\author{Ashish Acharya}
\email{aacha003@odu.edu}
\affiliation{\department{Department of Computer Science}\institution{Old Dominion University}\city{Norfolk}\state{Virginia}\country{USA}}
\author{Aman Al Amin}
\email{malam007@odu.edu}
\affiliation{\department{Department of Computer Science}\institution{Old Dominion University}\city{Norfolk}\state{Virginia}\country{USA}}
\author{Hong Qin}
\authornote{Corresponding author.}
\email{hqin@odu.edu}
\affiliation{\department{Department of Computer Science}\department{School of Data Science}\institution{Old Dominion University}\city{Norfolk}\state{Virginia}\country{USA}}

\maketitle

This supplement accompanies the main paper. Sections S1--S7 are cited there
by number, and S8 prints the prompt templates; section and equation numbers
without an ``S'' refer to the main paper. Evaluation results are generated
as described in Section~\mainref{sec:results}. Protocol
constants, the paid-API call and spend counts (S7), the development-only
trace (S5), and the development and calibration weight sweep (S6) are
transcribed from the run records. Hashes, digests, and run identifiers are
released with the artifact rather than printed here.

\section{Per-Family Results}
\label{sup:family}
Valid yield and invalid releases per requested output, each given as
numerator/denominator, by scenario family and policy. There is one table per
workflow, because conclusions are stated per workflow.
Tables~\ref{tab:sup-pt} and~\ref{tab:sup-sir} give the per-family cells.
Agentic cells are reported per model (Llama~3.3 70B, Gemma~3 27B, Mistral
Small 3.2 24B) where they differ; where the generator asserts identical
cells they share a column. The two sensitivity arms with the original
weights released no deliverable in any family
(Section~\ref{sup:deviations}).

\begin{table*}[h]
\caption{Published-table workflow, evaluation split. Each cell: valid yield (of requested); invalid releases; bold cells released at least one invalid deliverable. Agent columns: \Ag{0} and \Ag{2} give the same cells for all three models, and \Ag{4} the same for Llama and Mistral (L/M; G: Gemma; asserted by the generator).}
\label{tab:sup-pt}
\small
\begin{tabular}{@{}lccccccccc@{}}
\toprule
& \multicolumn{5}{c}{Deterministic policies} & \multicolumn{4}{c}{With agent (+Agent)} \\
\cmidrule(lr){2-6}\cmidrule(lr){7-10}
Family & \DR{0} & \DR{1} & \DR{2} & \DR{3} & \DR{4} & \DR{0} & \DR{2} & \DR{4} L/M & \DR{4} G \\
\midrule
Clean & 10/10; 0 & 10/10; 0 & 10/10; 0 & 0/10; 0 & 10/10; 0 & 10/10; 0 & 10/10; 0 & 10/10; 0 & 10/10; 0 \\
Benign & 10/10; 0 & 10/10; 0 & 10/10; 0 & 0/10; 0 & 10/10; 0 & 10/10; 0 & 10/10; 0 & 10/10; 0 & 10/10; 0 \\
Silent error & 2/10; 0 & 5/10; 0 & \textbf{5/10; 5} & 0/10; 0 & 9/10; 0 & 2/10; 0 & \textbf{5/10; 5} & 9/10; 0 & 6/10; 0 \\
Stale & 3/10; 0 & 5/10; 0 & 10/10; 0 & 2/10; 0 & 10/10; 0 & 3/10; 0 & 10/10; 0 & 10/10; 0 & 10/10; 0 \\
Shared ancestor & 0/10; 0 & 0/10; 0 & \textbf{0/10; 10} & 0/10; 0 & 10/10; 0 & 0/10; 0 & \textbf{0/10; 10} & 10/10; 0 & 8/10; 0 \\
Late invalidation & 1/10; 0 & 2/10; 0 & \textbf{7/10; 3} & 0/10; 0 & 2/10; 0 & 1/10; 0 & \textbf{7/10; 3} & 2/10; 0 & 2/10; 0 \\
\bottomrule
\end{tabular}
\end{table*}

\begin{table*}[h]
\caption{SIR workflow, evaluation split. Each cell: valid yield (of requested); invalid releases; bold cells released at least one invalid deliverable. Agent columns: \Ag{0} and \Ag{2} give the same cells for all three models, and \Ag{4} the same for Llama and Mistral (L/M; G: Gemma; asserted by the generator).}
\label{tab:sup-sir}
\small
\begin{tabular}{@{}lccccccccc@{}}
\toprule
& \multicolumn{5}{c}{Deterministic policies} & \multicolumn{4}{c}{With agent (+Agent)} \\
\cmidrule(lr){2-6}\cmidrule(lr){7-10}
Family & \DR{0} & \DR{1} & \DR{2} & \DR{3} & \DR{4} & \DR{0} & \DR{2} & \DR{4} L/M & \DR{4} G \\
\midrule
Clean & 10/10; 0 & 10/10; 0 & 10/10; 0 & 10/10; 0 & 10/10; 0 & 10/10; 0 & 10/10; 0 & 10/10; 0 & 10/10; 0 \\
Benign & 10/10; 0 & 10/10; 0 & 10/10; 0 & 10/10; 0 & 10/10; 0 & 10/10; 0 & 10/10; 0 & 10/10; 0 & 10/10; 0 \\
Silent error & 0/10; 0 & 5/10; 0 & \textbf{5/10; 5} & 7/10; 0 & 7/10; 0 & 0/10; 0 & \textbf{5/10; 5} & 7/10; 0 & 6/10; 0 \\
Stale & 0/10; 0 & 5/10; 0 & 10/10; 0 & 10/10; 0 & 10/10; 0 & 0/10; 0 & 10/10; 0 & 10/10; 0 & 10/10; 0 \\
Shared ancestor & 0/10; 0 & 0/10; 0 & \textbf{0/10; 10} & 10/10; 0 & 10/10; 0 & 0/10; 0 & \textbf{0/10; 10} & 10/10; 0 & 0/10; 0 \\
Late invalidation & 2/10; 0 & 2/10; 0 & \textbf{6/10; 4} & 2/10; 0 & 2/10; 0 & 2/10; 0 & \textbf{6/10; 4} & 2/10; 0 & 2/10; 0 \\
\bottomrule
\end{tabular}
\end{table*}

\paragraph{Execution integrity.} Table~\ref{tab:sup-execution} gives the
episode counts and final statuses of the evaluation runs: all \EpisodesAll\
planned episodes (\EvalPrimaryEpisodes\ deterministic,
\EvalSensEpisodes\ sensitivity, \LlmEpisodes\ agentic) ran to a recorded
final status with \InfraErrorsAll\ infrastructure errors. Every released
version has an oracle record, so no release is unlabeled. The audit
re-derived oracle labels on a sample of episodes without discrepancy,
reproduced all \AudPairedRows\ paired-difference rows, and replayed all
\AudLlmReplay\ agentic episodes byte-identically from their call logs
without a GPU. Of the ablation episodes, \AblTwoVBudget\ (\ValAll) and
\AblTwoMTwoBudget\ (\RandMatched) ended with the budget exhausted; they
are counted, not dropped.

\begin{table}[h]
\caption{Execution integrity of the evaluation runs (episode counts and final status, generated from the run records). Sensitivity: the two risk policies with their original weights (Section~\ref{sup:deviations}).}
\label{tab:sup-execution}
\footnotesize
\setlength{\tabcolsep}{3pt}
\begin{tabular}{@{}llrrrrrr@{}}
\toprule
Arm set & Arms & Ep. & Infra. & Compl. & Budget & Stop & Error \\
\midrule
Deterministic & 5 & 300 & 0 & 260 & 0 & 40 & 0 \\
Sensitivity & 2 & 120 & 0 & 120 & 0 & 0 & 0 \\
Agentic & 9 & 540 & 0 & 403 & 17 & 120 & 0 \\
\bottomrule
\end{tabular}
\par\smallskip\footnotesize Infra.: infrastructure errors. Final status: Compl.\ completed, Budget budget exhausted, Stop global stop (controller stopped the episode), Error runner error. Every episode ran to a recorded final status.
\end{table}

\paragraph{Calibration.} Table~\ref{tab:sup-calibration} gives the
calibration scores behind Section~\mainref{sec:results}, including the
original analysis and fill sensitivity values, and
Table~\ref{tab:sup-bins} the reliability bins.

\begin{table*}[h]
\caption{Calibration on the evaluation split. Post hoc (primary): risk of the latest consultation of each labeled version before release or withholding, usually after validator evidence. Fill sensitivity: the original analysis's risk where it exists, the post-hoc risk otherwise. Original: the original analysis (omits accepts; keeps the latest event's risk). Missing: versions without any consultation (external stale inputs, missing by design). Intervals: scenario-clustered bootstrap, 2000 replicates. On released subsets no version is invalid, so the reference Brier is 0 and skill is undefined.}
\label{tab:sup-calibration}
\footnotesize
\setlength{\tabcolsep}{3pt}
\begin{tabular}{@{}lccccccc@{}}
\toprule
Arm & $n$ (invalid) & Missing & Brier [95\% CI] & Ref. & Skill & ECE [95\% CI] & Fill / orig.\ Brier (orig.\ $n$, missing) \\
\midrule
\DR{3} & 327 (31) & 7 & 0.0153 [0.0108, 0.0218] & 0.086 & 0.82 & 0.078 [0.065, 0.092] & 0.110 / 0.122 (296, 38) \\
\DR{4} & 447 (42) & 10 & 0.0036 [0.0015, 0.0076] & 0.085 & 0.96 & 0.021 [0.015, 0.025] & 0.050 / 0.083 (255, 202) \\
\DRorig{3} & 60 (16) & 0 & 0.1956 [0.1491, 0.2498] & 0.196 & 0.00 & 0.001 & 0.196 / 0.196 (60, 0) \\
\DRorig{4} & 60 (16) & 0 & 0.1956 [0.1478, 0.2513] & 0.196 & 0.00 & 0.005 & 0.196 / 0.196 (60, 0) \\
\Ag{4} Gemma & 401 (74) & 10 & 0.0038 [0.0015, 0.0082] & 0.150 & 0.97 & 0.022 [0.016, 0.027] & 0.056 / 0.079 (268, 143) \\
\midrule
\multicolumn{8}{@{}l}{Released subsets: \DR{3} $n=55$, \DR{4} $n=106$, \Ag{4} Gemma $n=84$; 0 invalid in each.} \\
\multicolumn{8}{@{}l}{\Ag{4} with Llama, Mistral, and both API models: identical to \DR{4}.} \\
\bottomrule
\end{tabular}
\end{table*}

\begin{table}[h]
\caption{Post-hoc reliability bins, all labeled versions (non-empty bins only).}
\label{tab:sup-bins}
\small
\begin{tabular}{@{}lcrcc@{}}
\toprule
Arm & Bin & $n$ & Mean risk & Observed invalid \\
\midrule
\DR{3} & [0.0, 0.1) & 236 & 0.048 & 0.000 \\
\DR{3} & [0.1, 0.2) & 3 & 0.145 & 0.000 \\
\DR{3} & [0.2, 0.3) & 58 & 0.250 & 0.017 \\
\DR{3} & [0.9, 1.0] & 30 & 0.999 & 1.000 \\
\DR{4} & [0.0, 0.1) & 361 & 0.015 & 0.000 \\
\DR{4} & [0.1, 0.2) & 45 & 0.107 & 0.022 \\
\DR{4} & [0.9, 1.0] & 41 & 0.997 & 1.000 \\
\bottomrule
\end{tabular}
\end{table}

\paragraph{Agentic arms.} Table~\ref{tab:agentic} gives the per-model
agentic outcomes, calls, and tokens behind Section~\mainref{sec:results}.

\begin{table}[t]
\caption{Agentic arms, evaluation split (60 instances, 120 requested deliverables per arm). Calls and tokens are summed over \Ag{0}, \Ag{2}, and \Ag{4} and counted from the call logs. Every oracle-invalid approval was under \Ag{2}. $\Delta$ is per instance (two deliverables), with a case-group bootstrap 95\% interval.}
\label{tab:agentic}
\footnotesize
\setlength{\tabcolsep}{3pt}
\begin{tabular}{@{}lccc@{}}
\toprule
& Llama~3.3 & Gemma~3 & Mistral~S.~3.2 \\
\midrule
Calls, proposer / reviewer & 52 / 290 & 70 / 266 & 52 / 290 \\
Tokens, prompt / completion & 326.3k / 14.1k & 344.4k / 14.7k & 338.0k / 9.8k \\
Reviewer approvals & 290/290 & 266/266 & 290/290 \\
\quad of which oracle-invalid & 44 & 44 & 44 \\
Valid yield, \Ag{4} & 100/120 & 84/120 & 100/120 \\
Invalid/released, \Ag{2} & 37/120 & 37/120 & 37/120 \\
$\Delta$ valid, \Ag{4} vs.\ base & 0.00 & \ensuremath{-}0.27 & 0.00 \\
\quad 95\% interval & [0.00, 0.00] & [\ensuremath{-}0.44, \ensuremath{-}0.12] & [0.00, 0.00] \\
\bottomrule
\end{tabular}
\end{table}

\paragraph{Validation ablation.} Table~\ref{tab:ablation} gives the
post-hoc validation ablation behind Section~\mainref{sec:results},
including the defective first matched-rate arm \RandDefect\ and the paired
differences (disclosures in Section~\ref{sup:deviations}).

\begin{table}[t]
\caption{Post-hoc validation ablation, evaluation split (60 instances, 120 requested deliverables per arm). All three arms reuse the rules and recovery ladder of \DR{2}. Validators: mean inserted validator runs per instance; cost: mean contract cost per instance. Differences: ratio-of-sums (invalid/released, yield) or per-instance mean (cost), scenario bootstrap 95\% interval.}
\label{tab:ablation}
\footnotesize
\setlength{\tabcolsep}{3pt}
\begin{tabular}{@{}llcccc@{}}
\toprule
& & Invalid/ & Valid & Vali- & \\
Arm & Validation & released & yield & dators & Cost \\
\midrule
\DR{2} & rule, never fired & 37/120 & 83/120 & 0.00 & 11.07 \\
\ValAll & always & 0/119 & 112/120 & 5.98 & 25.17 \\
\RandDefect$^{\dagger}$ & random, low volume & 14/113 & 99/120 & 2.32 & 17.32 \\
\RandMatched & random, matched & 3/111 & 101/120 & 4.18 & 21.76 \\
\DR{4} & risk-targeted & 0/106 & 100/120 & 4.13 & 19.65 \\
\midrule
\multicolumn{6}{@{}l}{\DR{4}$-$\RandMatched: invalid/released \ensuremath{-}0.027 [\ensuremath{-}0.063, 0.000],} \\
\multicolumn{6}{@{}l}{\quad not resolved at $n=60$; yield \ensuremath{-}0.008 [\ensuremath{-}0.075, 0.058];} \\
\multicolumn{6}{@{}l}{\quad cost \ensuremath{-}2.12 [\ensuremath{-}3.86, \ensuremath{-}0.38].} \\
\multicolumn{6}{@{}l}{\DR{4}$-$\ValAll: invalid/released 0.000 [0.000, 0.000];} \\
\multicolumn{6}{@{}l}{\quad yield \ensuremath{-}0.100 [\ensuremath{-}0.167, \ensuremath{-}0.042]; cost \ensuremath{-}5.52 [\ensuremath{-}7.50, \ensuremath{-}3.91].} \\
\bottomrule
\end{tabular}
\par\smallskip\footnotesize\raggedright $^{\dagger}$Defective specification (Supplement~S6): the matched rate was defined per candidate consultation but applied only where a validator was admissible, so \RandDefect\ made 139 validator insertions against \DR{4}'s 248. It is reported for completeness and is not evidence about targeting. \RandMatched\ corrects the rate (251 insertions, within the pre-set 211--285 band). At matched volume the safety difference is not resolved at $n=60$ (targeting not shown); \DR{4} was cheaper.
\end{table}

\FloatBarrier
\section{Proposal Menus and Prompts}
\label{sup:prompts}
Each menu option is one recovery action (retry, select an approved
alternative, repair, or invalidate and recompute) applied to the node under
decision. Options carry no scenario identifiers, family names, fault
labels, or oracle fields. The menus and prompts are released as files with
the artifact.

\begin{table}[t]
\caption{Proposal menus, as fixed before the agentic evaluation run. ``Any'': every node of the workflow.}
\label{tab:sup-menus}
\scriptsize
\setlength{\tabcolsep}{2.5pt}
\begin{tabularx}{\linewidth}{@{}>{\raggedright}p{1.7cm}>{\raggedright}p{1.15cm}>{\raggedright}p{1.32cm}Y@{}}
\toprule
Option & Action & Nodes & Description \\
\midrule
\multicolumn{4}{@{}l}{\emph{Published table workflow}} \\
\texttt{retry} & retry & any & Run the node's current operation again on the same inputs, with no changes. \\
\texttt{repair} & repair & any & Run the node's current operation again with the previous attempt's check results attached as diagnostics. \\
\texttt{recompute} & invalidate, recompute & any & Invalidate the current version and recompute the node from its current inputs, applying any input updates published during the run. \\
\texttt{table\_\allowbreak{}from\_\allowbreak{}verified\_\allowbreak{}transcription} & select alternative & \texttt{source\_table} & Rebuild the source table from the frozen verified transcription of the publication instead of the runtime table text. \\
\texttt{means\_\allowbreak{}decimal} & select alternative & \texttt{a\_means} & Recompute the per-padding means with exact decimal arithmetic instead of floating point. \\
\midrule
\multicolumn{4}{@{}l}{\emph{SIR workflow}} \\
\texttt{retry} & retry & any & Run the node's current operation again on the same inputs, with no changes. \\
\texttt{repair} & repair & any & Run the node's current operation again with the previous attempt's check results attached as diagnostics. \\
\texttt{recompute} & invalidate, recompute & any & Invalidate the current version and recompute the node from its current inputs, applying any input updates published during the run. \\
\texttt{spec\_\allowbreak{}from\_\allowbreak{}registry} & select alternative & \texttt{model\_spec} & Build the model specification from the registry parameters in the runtime inputs instead of the working configuration. \\
\texttt{solver\_\allowbreak{}rk45\_\allowbreak{}tight} & select alternative & \texttt{trajectory\_a}, \texttt{trajectory\_b} & Re-solve the ODE system with RK45 at rtol 1e-11 and atol 1e-13. \\
\texttt{solver\_\allowbreak{}lsoda} & select alternative & \texttt{trajectory\_a}, \texttt{trajectory\_b} & Re-solve the ODE system with LSODA at rtol 1e-11 and atol 1e-13. \\
\texttt{solver\_\allowbreak{}dop853} & select alternative & \texttt{trajectory\_a}, \texttt{trajectory\_b} & Re-solve the ODE system with DOP853 at rtol 1e-11 and atol 1e-13. \\
\bottomrule
\end{tabularx}
\end{table}

\paragraph{Model settings.} Ollama's native chat API, temperature \LlmTemperature, seed \LlmSeed, a context of \LlmNumCtx\ tokens, at most \LlmNumPredict\ output tokens, a timeout of \LlmTimeout\,s, and no retries. All three models use Ollama's default \texttt{\LlmQuant} quantization and are pinned by manifest digest. Every call log entry carries a hash over the four prompt templates and the two JSON response schemas, and verification refuses any entry with another hash; the hashes and digests are in the artifact.

\paragraph{Proposal menus.} Table~\ref{tab:sup-menus} lists every option. An option applies only to the node under decision and only at the nodes listed. Declared parameters (for example a solver's tolerances) are recorded with the call but are not passed to the operation; a solver setting is executed only through the named, contract-approved alternative.

\paragraph{Prompts.} The proposer and reviewer prompts are printed in full in Section~\ref{sup:templates}. The development pilot and the calibration sanity run used an earlier reviewer prompt, replaced by the single development-only revision (Section~\ref{sup:deviations}) before any evaluation episode ran; the proposer templates are unchanged. The earlier templates were never committed to the repository. They were recovered from the logged command that wrote them and verified two ways: with the proposer templates they reproduce the earlier prompt hash exactly, and the system prompt equals the one recorded in the pilot call logs. They are also printed in Section~\ref{sup:templates}.

\FloatBarrier
\section{Risk Model Details}
\label{sup:features}
\paragraph{Features.} Features are computed by the controller's declared
feature functions. A check summary over a set of check records has five
fields: the number of checks, failed checks, optional checks run, the
maximum absolute check value, and an indicator that a value is available;
unknown outcomes are counted separately (Table~\ref{tab:sup-features}).
``Constant'' marks features with a single value on all 2000 development
rows. Ancestors are taken along scientific-input edges only, one entry per
distinct ancestor version; diagnostic edges are not traversed.

\begin{table}[t]
\caption{Risk features. L: local (\DR{3}, \DR{4}); A: ancestral
(\DR{4} only).}
\label{tab:sup-features}
\small
\begin{tabularx}{\linewidth}{@{}lcYc@{}}
\toprule
Feature & Set & Meaning & Constant \\
\midrule
\texttt{n\_checks} & L & checks recorded on the version & \\
\texttt{n\_fail} & L & failed checks & \\
\texttt{n\_unknown} & L & checks with unknown outcome & yes \\
\texttt{n\_optional} & L & optional checks run & \\
\texttt{max\_value} & L & maximum absolute check value & \\
\texttt{has\_value} & L & a check value is available & \\
\texttt{attempts\_used} & L & attempts used on the node & \\
\texttt{is\_recovery\_attempt} & L & produced by a recovery action & \\
\texttt{is\_deliverable} & L & node is a deliverable & \\
\texttt{infra\_error} & L & last attempt had an infrastructure error & yes \\
\texttt{optional\_available} & L & optional checks still available & \\
\texttt{issue\_check} & L & open check issues & \\
\texttt{issue\_execution} & L & open execution issues & yes \\
\texttt{issue\_version} & L & open version issues & \\
\texttt{issue\_integrity} & L & open integrity issues & yes \\
\texttt{issue\_late} & L & open late-evidence issues & \\
\texttt{issue\_closure} & L & open closure issues & yes \\
\texttt{issue\_input} & L & open input issues & yes \\
\midrule
\texttt{anc\_n\_checks} & A & checks over all ancestors & \\
\texttt{anc\_n\_fail} & A & failed ancestral checks & yes \\
\texttt{anc\_n\_unknown} & A & unknown ancestral checks & yes \\
\texttt{anc\_n\_optional} & A & optional ancestral checks run & \\
\texttt{anc\_max\_value} & A & maximum absolute ancestral check value & \\
\texttt{anc\_has\_value} & A & an ancestral value is available & \\
\texttt{anc\_count} & A & distinct ancestor versions & \\
\texttt{anc\_recovered} & A & ancestors produced by recovery & \\
\texttt{anc\_resolved\_issues} & A & distinct issue causes in the lineage, each counted once & yes \\
\bottomrule
\end{tabularx}
\end{table}

\paragraph{Action benefits.} Fix, break, sensitivity, and false-alarm rates
are estimated per (action, context) from development action records, with
a pooled fallback (Table~\ref{tab:benefit}). One rule, fixed before the
fit, covers every action. An entry needs at least five invalid-before
records with a labeled outcome; otherwise the action is not scored and the
policy abstains rather than take it. With fewer than five valid-before
records, the valid-before rate takes its worst case: $b_a=1$ for corrective
actions, and false alarm equal to sensitivity (an uninformative validator).
Because the loss of Eq.~(\mainref{eq:action}) is non-decreasing in $b_a$,
this bound can only lower the benefit of recovery, biasing against \DR{3}
and \DR{4}. Every corrective entry uses the bound: almost all labeled
pre-action versions at blocked or invalidated consultations were invalid.
Pooled entries are repair (fix $0.146$; 48/0), select alternative ($0.889$;
9/0), and invalidate-and-recompute ($1.0$; 51/2). Repair and select
alternative at version-blocked states lack support (3 and 1 records). The
validator's measured cost is $1.98$. Table and weights are identical for
\DR{3} and \DR{4}, which differ only in features; this matched pair was
checked before the models were fixed.

\begin{table}[t]
\caption{Action-benefit table (development records only). Entries
are action, situation, mandatory-fail status, and issue kind. ``Bound'':
worst-case valid-before rate (fewer than five valid-before records).}
\label{tab:benefit}
\small
\begin{tabularx}{\linewidth}{@{}Yrrrl@{}}
\toprule
Entry & Rate & Inv. & Val. & Source \\
\midrule
validator, candidate, no fail & se $0.842$, fa $0.000$ & 38 & 110 & estimated \\
repair, blocked, fail, check & fix $0.089$ & 45 & 0 & bound \\
select alt., blocked, fail, check & fix $0.875$ & 8 & 0 & bound \\
recompute, blocked, no fail, version & fix $1.0$ & 36 & 0 & bound \\
recompute, invalidated, no fail, late & fix $1.0$ & 15 & 2 & bound \\
\bottomrule
\end{tabularx}
\end{table}

\paragraph{Pre-evaluation choice diagnostic.} Replaying the calibration
consultations without budget limits, \DR{3} and \DR{4} choose the
validator at 351 and 340 of 459 candidates, accept 49 and 60, and abstain on
59 each; they choose invalidate-and-recompute at all 10 blocked and all 8
invalidated consultations.

\FloatBarrier
\section{Oracle Definitions}
\label{sup:oracles}
\paragraph{Verdicts.} Each candidate version receives one of three verdicts.
It is \emph{unobserved} if there is no candidate (missing version or value),
no binding for the deliverable, or the reference does not apply (parameters
outside the analytical domain, or a horizon too short for an asymptotic
target). Unobserved records are never counted as valid or invalid. It is
\emph{invalid} if a candidate exists and is malformed, non-finite, or outside
the contract tolerance of the reference, and \emph{valid} if a candidate
exists and agrees with the reference. Intermediate fields borrow the
tolerance of the deliverable field they feed. The oracle never runs
production code and never derives a reference from the candidate, runtime
tables, or hidden expected answers.

\paragraph{Published table.} References are exact decimal computations
(50-digit precision) on the oracle's own hash-verified copy of the source
table. No production routine is imported. Group means are exact. Rankings
treat exact decimal equality on the published three-decimal values as ties,
and tied conditions may appear in any order.

\paragraph{SIR model.} With populations normalized by $N$,
$\dot s=-\beta s i$, $\dot i=\beta s i-\gamma i$, $\dot r=\gamma i$, and
$R_0=\beta/\gamma$. No relation below integrates the ODE.
\begin{itemize}
\item Phase-plane invariant: $i(s)=i_0+s_0-s+R_0^{-1}\ln(s/s_0)$.
\item Final size: $s_\infty$ is the unique root of $i(s)=0$ in the
interval $(0,\min(s_0,1/R_0)]$. It is solved by bisection in $\log s$ and
cross-checked against the Lambert-W closed form; $r_\infty=1-s_\infty$.
\item Peak: if $R_0 s_0>1$, $i_{\max}=i_0+s_0-R_0^{-1}(1+\ln R_0 s_0)$ at
$s=1/R_0$; otherwise $i_{\max}=i_0$ at $t=0$.
\item Time map: $t(s)=\int_{\ln s}^{\ln s_0}\mathrm{d}u/(\beta\, i(e^u))$ by
adaptive Gauss--Legendre quadrature. It is inverted by bisection to give
$s$, $i$, and $r$ at a finite horizon.
\end{itemize}
Asymptotic targets such as final size and an interior peak are observable
only if the horizon is long enough. Otherwise the field is unobserved, never
guessed.

\FloatBarrier
\section{Worked Recovery Trace (Development Scenario)}
\label{sup:trace}
\begingroup\emergencystretch=3em
This is a development late-invalidation episode under \DR{2}, taken from
the development pilot of the evaluation arm set; it is never an evaluation
episode. Its manifest and decision record are released with the artifact,
and verification re-checks the manifest digest held in the decision record.
\begin{itemize}
\item \emph{Scenario.} Published-table workflow, family late invalidation.
The working copy of the source table swaps the 50 and 100\,bp padding
labels of one condition (BWA-MEM with GATK HaplotypeCaller). No mandatory
check can see this. An erratum with the corrected table arrives after
\texttt{a\_report} has run.
\item \emph{Before the issue.} All seven nodes ran on the corrupted copy and
passed their mandatory checks: \texttt{source\_table.v1}, then
\texttt{b\_rows}, \texttt{b\_ranking}, \texttt{b\_report} (v2--v4) and
\texttt{a\_rows}, \texttt{a\_means}, \texttt{a\_report} (v5--v7). The
unreleased \texttt{a\_report.v7} gave 50\,bp as the best padding (mean MCC
0.284 at 50\,bp, 0.247 at 100\,bp).
\item \emph{Issue raised.} The erratum opens a late-evidence issue on
\texttt{source\_table.v1} and on all six descendants built from it. The
situation is ``invalidated'', and the admissible actions are abstain,
global stop, and invalidate-and-recompute.
\item \emph{Decision and risk.} The fixed rule of \DR{2} chose
invalidate-and-recompute at \texttt{source\_table}. \DR{2} uses no risk
model, so none was recorded.
\item \emph{Graph change.} Revision 0 to 1 adds one edge, from the
recompute attempt to \texttt{source\_table}, to the six-edge graph. No edge
is removed.
\item \emph{Recomputed descendants.} \texttt{source\_table.v8} was rebuilt
from the corrected table. The six descendants were then recomputed in the
scenario's branch order as v9--v14 (\texttt{b\_rows}, \texttt{b\_ranking},
\texttt{b\_report}, \texttt{a\_rows}, \texttt{a\_means},
\texttt{a\_report}). Versions v1--v7 were superseded. The recomputation cost 10.25, the same as the first pass, for
a total of 20.5 against a budget of 60.
\item \emph{Outcome.} Release closure was verified, and \texttt{a\_report.v14}
and \texttt{b\_report.v11} were released. Nothing was withheld or retracted.
The best padding is now 100\,bp (0.269 against 0.262), and in the ranking
the 50 and 100\,bp variants of the affected condition trade ranks 9 and 12. The pilot used a
stub oracle, so no validity verdict was recorded. Recomputing the padding
means directly from the corrected table gives exactly the released values.
\end{itemize}
\endgroup

\FloatBarrier
\section{Deviation Log}
\label{sup:deviations}
Every change below to the protocol fixed before execution, except the four
marked \emph{post hoc}, was decided and fixed before any evaluation oracle output
was computed or opened. The changes marked \emph{observed}
were made after development or calibration behavior had been seen; we
disclose them as selection on non-evaluation data.

\begin{enumerate}
\item \textbf{Candidate decisions by loss, not thresholds.} \DR{3} and
\DR{4} originally had fixed validate and abstain thresholds. They now decide
at candidates through Eq.~(\mainref{eq:action}). \DR{2} keeps its fixed rule.
Fixed before any fit, together with the fitting, calibration, and
terminology thresholds.
\item \textbf{Candidate-point exploration} (observed). The development
exploration policy first explored only non-candidate points (29 of 913
consultations). Because \DR{2} never inserted a validator, there were no
validator outcome records. The policy now also inserts the validator with
probability $\varepsilon$ where \DR{2} would accept. \DR{2} itself was not
changed, and the development trajectories were regenerated. An earlier
development run had been refused for code drift and was not used.
\item \textbf{Support rule and worst-case bound.} Corrective actions at
blocked states could never reach valid-before support. Rather than
inventing a prior, every action now uses the support rule and worst-case
bound of Section~\mainref{sec:risk}. Adopted before any fit. Applied to the
loss as first formulated, the bound could raise benefit. A correction
charging a broken valid version made the loss monotone; the next item
supersedes it.
\item \textbf{Loss accounting and weight scale} (observed). The original
models used $\lambda_I=10$, $\lambda_W=2$, the scaffold's test
constants, with no non-delivery term. With these weights the criterion reduced to
``accept iff $p<1/6$, else abstain''. On the calibration split \DR{4}
abstained on 417 of 477 consultations and \DR{3} on 428; validators and
corrective actions were never chosen. The accounting was aligned with the
predefined valid-yield outcome, and the weights were anchored to the
contract budget (Section~\mainref{sec:risk}); the 5:1 harm ratio and $p^\ast$
are unchanged. The result was fixed as an intermediate version.
\item \textbf{Risk rows from all situations} (observed). The intermediate
models were fitted on candidate rows only, so mandatory-failure features were
effectively constant. At blocked states they gave $p\approx0.25$--$0.36$
although every labeled blocked version was invalid. On the calibration split,
their candidate choices were the same as those of the final models,
but both abstained at all 10 blocked
and all 8 invalidated consultations. On development, both chose 106
abstentions and 8 alternatives at blocked states. Refitting on all labeled
situations with features, estimator, table, and weights unchanged gave the
final models, used for evaluation. This was the last pre-evaluation change.
\textbf{The intermediate version is superseded and was not run on
evaluation.}
\item \textbf{Label semantics.} Labels were kept as oracle output for the 9
intermediate versions with an unconstrained solver setting (Section~\mainref{sec:design}).
No cause label was created, and the contracts were not changed.
\item \textbf{Agentic arms} (added after primary execution). The
proposer/reviewer arms (Section~\mainref{sec:agentic}) were added by author
decision after the primary evaluation episodes had run and before any
evaluation oracle output was computed or opened. They run under their own
protocol, fixed before their run, and are never pooled with the
deterministic arms, whose analyses are unchanged.
\item \textbf{Reviewer-prompt revision} (observed, development only;
pilot note). Under the first reviewer prompt, \Ag{2} made 0 releases in the
development pilot: the reviewer returned ``uncertain'' because optional
validators had not run, and at that state the menu offers only repair. One
generic revision followed (Section~\mainref{sec:agentic}), the pilot was rerun,
and the prompt was fixed with that result whatever it showed. Adding a
validator option to the menu was rejected, because it would give \Ag{2}
\DR{4}'s action space. Calibration data served only as a sanity check.
\item \textbf{Calibration analysis} (post hoc, after evaluation). The
original analysis has two biases. It took predicted risk only from decisions
that targeted a version; accepts record none, so low-risk accepts were
missing from the calibration pairs. It also kept the risk of the latest
such decision, often the consultation before a validator ran, so the risk
it scored was stale. The evaluated code was not changed. A separate post-hoc
module, written and run by the automated auditor (a separate agent, read-only
on the runs), pairs each labeled
version with the risk of its latest consultation in the digest-verified
decision record (scenario-clustered bootstrap, 2000 replicates). It
reproduces the original values exactly when restricted to the original
pairing, is the source for all calibration claims, and the original values
are kept (Table~\ref{tab:sup-calibration}). Versions never consulted
(external stale inputs) remain unpaired by design.
\item \textbf{Paid-API addendum} (post hoc, after the agentic results had
been seen). By author decision, \Ag{0}, \Ag{2}, and \Ag{4} were rerun
with two paid models on Vertex AI under a new protocol fixed before the run:
360 evaluation episodes after a 54-episode development pilot, with a spend
cap of USD~10. It is not part of the planned study, is never pooled, and is
reported only in Supplement~S7. Further API runs are limited to development
and calibration data. Adding the API client changed four files whose hashes
had been recorded for the open-weight agentic run, after that run had
finished. Its results are unaffected: the original files were rebuilt,
verified against the recorded hashes, and archived with the artifact, so
its replays use the original code.
\item \textbf{Validation ablation} (post hoc, after evaluation; prompted by
the \DR{2} policy's validator trigger never firing). Three arms reuse that policy's rules and
recovery ladder on the same \ResInstances\ evaluation instances, budgets,
and oracle, with new code only and no hashed file changed, a development
smoke run, and a fixed specification before each evaluation run
(\AblEpisodes\ episodes, no model calls). \ValAll\ inserts every available
optional validator wherever insertion is admissible. \RandDefect\ was meant
to validate
at random at \DR{4}'s observed rate, an evaluation statistic used by design
as a matched control, but the rate was defined per candidate consultation
and applied only where insertion was admissible, so it made
\AblTwoMValidators\ insertions against \DR{4}'s \AblFourInsertions. This
specification error is ours; \RandDefect\ is reported for completeness
(Table~\ref{tab:ablation}) and is not evidence about targeting. \RandMatched\
applies the corrected per-admissible rate, fixed anew before its run, and made
\AblTwoMTwoValidators\ insertions, inside the pre-set
\AblVolumeLo--\AblVolumeHi\ band. Its reading rule (targeting shown only if
\RandMatched\ released an invalid deliverable and the interval for \DR{4}$-$\RandMatched\
in invalid per released lay entirely below $0$) was fixed while the run was
in progress, before any outcome was computed or inspected; the outcome is
``targeting not shown''.
\item \textbf{No-release-control reference} (post hoc, a descriptive
reference, not a policy arm; prompted by the question ``compared with
what?''). For each of the 120 eligible evaluation deliverables it takes the
first version produced by execution in the \DR{0} episode, before any check,
block, or recovery could act on it, and labels it with the oracle.
The definition was pinned before any computation and amended, also before
computation, to report bounds because \DR{0} stops at its first mandatory
blocker (\NoHarnNotProdNum/\NoHarnNotProdDen\ deliverables never produced).
The stored-label source for the point value was chosen after a feasibility
pass had displayed per-category counts; no rule was changed. The generator's
no-intervention labels agreed with the oracle on all
\NoHarnAgreeN\ deliverables whose first version no recovery action shaped
in any of the five deterministic policies; the other \NoHarnUncheckable\ could not be
cross-checked and are all labeled affected. First-produced content was
identical across arms wherever compared (\NoHarnHashAllArms/120 deliverables
produced in all five arms; no mismatch pairwise). No episode, model call, or
oracle run was added. Comparisons with policy arms are descriptive, without
paired intervals.
\ifdefined\arxivbuild\else
\item \textbf{Manuscript.} The column mapping policies to an earlier
protocol ladder was removed for anonymity; policy definitions are unchanged.
\fi
\end{enumerate}

\paragraph{Known tool defects.} Found after the runs; the evaluated code was
not changed. (i) The original analysis took predicted risk only from
decisions that targeted a version, so accepts, which record none, were
missing from the calibration pairs, and the risk kept was often a
pre-validator one; calibration claims use the post-hoc recomputation
(item 9). (ii) The agentic runner's summaries read a token
field that the call-log summary does not emit, so they record zero tokens;
the paper's token counts come from the call logs. (iii) The release-closure
code can construct a dependency without a version, and crash, when an
upstream offender is withheld. No reported run reached this path: no
episode recorded a closure decision or a controller error, while a
reproduction still crashes. (iv) Verification does not re-check the
record of pinned hashes itself, and episode status, arm role, case group, and the
hidden labels lie outside the hashed files.

\paragraph{Sensitivity arms.} \DRorig{3} and \DRorig{4} run the original
models (item 4) with $\lambda_I=10$, $\lambda_W=2$, and the original
accounting on the evaluation split. They were pre-declared when the
weights were changed and add 120 episodes,
reported here only. Both released nothing: valid yield
\ResThreeVoneValid/\ResThreeVoneReq\ and \ResFourVoneValid/\ResFourVoneReq, and all
\ResThreeVoneFBden\ clean and benign deliverables withheld, at a mean cost of
\ResThreeVoneCost\ per instance (episode counts in
Table~\ref{tab:sup-execution}). Their risk is constant on evaluation (Brier
\CalThreeVoneBrier\ and \CalFourVoneBrier\ against a reference of
\CalVoneRef; Table~\ref{tab:sup-calibration}).

\paragraph{Weight sweep (final models, development and calibration only).}
Written with the final models, before any evaluation episode, and never
used to choose weights, the sweep re-scores the one-step choice of the
final \DR{3} and \DR{4} models at every recorded consultation (477 calibration,
2000 development) on a grid $\lambda_W=s\cdot 30$, $s\in\{0.25,0.5,1,2,4\}$,
and $\lambda_I=r\lambda_W$, $r\in\{2,5,10,20\}$. It is not part of the
audited evaluation outputs. The harm ratio $r$ matters far more than the
scale $s$. On calibration at the chosen point ($s=1$, $r=5$), \DR{3}
and \DR{4} release at 49 and 60 consultations, insert a validator at 351
and 340, and abstain at 59. At $r=2$ neither abstains at any scale (107 to
119 releases). At $r=20$ they abstain at 410 and 399 of 477 at every
scale. The 18 blocked and invalidated calibration consultations go to
invalidate-and-recompute at every grid point. Development shows the same
pattern (at the chosen point 304 and 271 abstentions of 2000; at $r=20$
about 1550 to 1780 at every scale).

\FloatBarrier
\section{Post-Hoc Paid-API Addendum}
\label{sup:api}
This addendum is a disclosed deviation (Section~\ref{sup:deviations}): it
was requested by the authors after the agentic results of
Section~\mainref{sec:agentic} had been seen, and it is not part of the
planned study. Its episodes are paired with the deterministic evaluation by
instance and seed but are never pooled with any other arm. The main paper
cites it only as consistent with the open-weight results; no conclusion
rests on it.

\begin{table}[h]
\caption{Post-hoc paid-API addendum (not pooled). Valid yield of 120; calls proposer/reviewer and tokens prompt/completion from the call logs, excluding Gemini's reasoning tokens (69.1k in total, billed as output); all reviewer verdicts were approvals.}
\label{tab:sup-api}
\footnotesize
\setlength{\tabcolsep}{3pt}
\begin{tabular}{@{}llcccc@{}}
\toprule
Model & Arm & Valid & Inv./rel. & Calls & Tokens \\
\midrule
Grok~4.3 & \Ag{0} & 48 & 0/48 & 0/48 & 63.0k/2.2k \\
 & \Ag{2} & 83 & 37/120 & 16/128 & 184.8k/7.0k \\
 & \Ag{4} & 100 & 0/106 & 36/114 & 191.3k/7.8k \\
Gemini~3.6 Flash & \Ag{0} & 48 & 0/48 & 0/48 & 55.0k/1.7k \\
 & \Ag{2} & 83 & 37/120 & 16/128 & 160.8k/5.4k \\
 & \Ag{4} & 100 & 0/106 & 36/114 & 166.6k/6.0k \\
\bottomrule
\end{tabular}
\end{table}

\paragraph{Protocol.} The arms \Ag{0}, \Ag{2}, and \Ag{4} of
Section~\mainref{sec:agentic} were run with two commercial models served
through Vertex AI (\texttt{generateContent}, global endpoint), Grok~4.3 and
Gemini~3.6 Flash, giving 360 evaluation episodes under a separate protocol
fixed before the run. Prompts, menus, roles, and controller semantics are
those of the open-weight arms. Decoding uses temperature $0$, the same seed,
and a JSON response schema. The output limit is 4{,}096 tokens, not 512,
because Gemini's reasoning tokens count against it. Reasoning was left at
each model's default. A model reply is never repaired or re-asked, as in the
open-weight arms. Unlike them, a request that received \emph{no} reply
(HTTP 429 or 5xx, or a connection error) could be re-sent up to four times
with backoff, and every attempt is logged. A 54-episode development pilot
preceded the evaluation run. The spend cap was USD~10; the pilot and the
evaluation run cost about USD~0.12 and USD~1.19 at list price, including
Gemini's reasoning tokens. The models are pinned by a hash of the endpoint
identity (project, location, publisher, model), not by weights. Each call
log records the response identifier and, where the provider reports it, the
served model version. Gemini reported \texttt{gemini-3.6-flash}; Grok
reported none. Because a provider can change the model behind a name, the
runs can be reproduced only by replaying the call logs. Further API calls
are restricted to development and calibration data.

\paragraph{Results.} Table~\ref{tab:sup-api} gives the addendum outcomes.
Both API models produced, in all \AudApiEpisodes\ episodes, the same final
releases and trajectories as Llama and Mistral in the open-weight arms, and
an automated audit by a separate read-only agent replayed all
\AudApiReplay\ episodes from their call
logs. The reviewer approved all \RevApiN\ candidates it saw, including
\RevApiApproveInvalid\ oracle-invalid ones; as in the main paper, this is
structural and carries no safety evidence. These results are never pooled
with any other arm. All \AudApiCalls\ calls returned a parseable, on-menu
reply, with \AudApiErrors\ errors. Grok was rate-limited
(HTTP 429) on 45 of its 342 calls; every one succeeded on a re-send, but 3
calls needed all five allowed attempts, so one more refusal would have
become an error and an abstention. Gemini needed no
re-sends. Agreement with the open-weight arms holds at the level of
individual decisions: every proposer choice and reviewer verdict equals
Llama's and Mistral's, for each model and arm (180 of 180 episodes per
model). Gemma differs in 36 of 180.

\FloatBarrier
\clearpage
\onecolumn
\section{Prompt Templates}
\label{sup:templates}

The four templates used in every agentic evaluation episode, followed by the two earlier reviewer templates used only in the development pilot and the calibration sanity run. Each file is reproduced verbatim and matches the hash recorded before the evaluation run. Lines are wrapped for printing; the arrow marks a continuation.

\subsection*{Proposer, system prompt}
\begin{lstlisting}
You are the corrective-action proposer inside a provenance-tracking scientific workflow controller.

The controller has already decided that the current artifact must be corrected rather than accepted, validated, withheld or stopped. Your only task is to choose exactly ONE corrective option from the menu you are given.

Action semantics (fixed by the task contract):
- retry: re-run the same operation unchanged; appropriate for transient infrastructure errors.
- select_alternative: run a contract-approved alternative method for this artifact.
- repair: re-derive the artifact from its approved inputs using the declared repair procedure.
- invalidate_recompute: discard the current version and recompute it from current inputs (applies pending source updates).

Rules:
- Use only the evidence shown: the contract, check results, open issues, the lineage summary, and the remaining budgets.
- Prefer the option whose declared effect addresses the open issues; consider remaining attempts, repairs and cost.
- You cannot release, accept or skip; every correction is re-checked by the controller before any use.
- Reply with a single JSON object matching the schema: {"option_id": <one of the listed ids>, "rationale": <one or two sentences>}. No other text.
\end{lstlisting}
\subsection*{Proposer, user prompt template}
\begin{lstlisting}
## Task contract
Workflow: {{ contract.workflow }}
Scientific target: {{ contract.scientific_target }}
Evidence tier: {{ contract.evidence_tier }}
Deliverables: {{ contract.deliverables | join(", ") }}
Declared checks for this artifact:
{% for c in contract.node_checks %}- {{ c.name }} ({{ "mandatory" if c.mandatory else "optional" }}, cost {{ c.cost }})
{% else %}- none
{% endfor %}
## Artifact needing correction
Artifact: {{ obs.logical_id }}{{ " (deliverable)" if obs.is_deliverable else "" }}
Situation: {{ obs.situation }}
Produced by: {{ obs.produced_by }}{% if obs.last_outcome %}; last attempt outcome: {{ obs.last_outcome }}{% endif %}

Check results on this version:
{% for c in obs.checks %}- {{ c.name }}: {{ c.outcome }}{{ " (mandatory)" if c.mandatory else "" }}{% if c.value is not none %}, normalised value {{ c.value }}{% endif %}{% if c.message %}; {{ c.message }}{% endif %}
{% else %}- none recorded
{% endfor %}
Open issues:
{% for i in obs.issues %}- [{{ i.kind }}] {{ i.description }}
{% else %}- {{ obs.issue_kinds | join(", ") if obs.issue_kinds else "none" }}
{% endfor %}
Lineage (scientific-input ancestors):
{% for a in obs.lineage %}- {{ a.logical_id }}: produced by {{ a.produced_by }}; checks {{ a.n_pass }} pass / {{ a.n_fail }} fail / {{ a.n_unknown }} unknown; resolved issues {{ a.resolved }}
{% else %}- none
{% endfor %}
## Budgets
Attempts on this artifact: {{ obs.attempts_used }} of {{ obs.max_attempts }}
Repairs remaining (episode): {{ obs.repairs_remaining }}
Cost remaining (episode): {{ obs.cost_remaining }}

## Menu (choose exactly one option_id)
{% for o in options %}- {{ o.option_id }}: {{ o.action }}{% if o.alternative %} [{{ o.alternative }}]{% endif %} -- {{ o.description }}{% if o.params %} (params: {{ o.params | tojson }}){% endif %}
{% endfor %}
\end{lstlisting}
\subsection*{Reviewer, system prompt}
\begin{lstlisting}
You are the release reviewer inside a provenance-tracking scientific workflow controller.

A deliverable artifact has passed every blocking check and the controller proposes to accept it for release. You must decide whether the recorded evidence supports releasing it.

The task contract divides the checks on an artifact into two kinds:
- Required checks: the contract marks them mandatory; they are the contract's release criteria for this artifact.
- Optional checks: extra validators the contract offers at additional cost but does not require for release. They may or may not have been run.

Rules:
- Use only the evidence shown: the contract, the required and optional check results, open issues, the lineage summary, and the remaining budgets.
- Judge the artifact on the evidence that was recorded. Read check values and messages, and the lineage, for anything that contradicts the contract.
- An optional check that has not been run is not, by itself, grounds for uncertain or reject. Do not ask for checks the contract does not require.
- An optional check that was run is evidence: weigh its result like any other evidence.
- approve: every required check passed and the recorded evidence supports release.
- reject: the evidence indicates the artifact is likely wrong or not what the contract asks for (for example a failed check, or values or lineage results that contradict the contract).
- uncertain: reserve this for a required check that is missing, not run or inconclusive, or for evidence that is contradictory.
- Anything other than approve blocks release of this version; the controller may then correct the artifact or withhold it.
- Reply with a single JSON object matching the schema: {"verdict": "approve" | "reject" | "uncertain", "reason": <one or two sentences>}. No other text.
\end{lstlisting}
\subsection*{Reviewer, user prompt template}
\begin{lstlisting}
## Task contract
Workflow: {{ contract.workflow }}
Scientific target: {{ contract.scientific_target }}
Evidence tier: {{ contract.evidence_tier }}
Deliverables: {{ contract.deliverables | join(", ") }}
Declared tolerances:
{% for name, t in contract.tolerances %}- {{ name }}: abs {{ t.abs_tol }}, rel {{ t.rel_tol }}{% if t.unit %} ({{ t.unit }}){% endif %}
{% else %}- none
{% endfor %}
## Release candidate
Artifact: {{ obs.logical_id }} (deliverable)
Produced by: {{ obs.produced_by }}{% if obs.last_outcome %}; last attempt outcome: {{ obs.last_outcome }}{% endif %}

Required checks for this artifact (contract release criteria):
{% for c in review.required %}- {{ c.name }}: {{ c.status }}{% if c.value is not none %}, normalised value {{ c.value }}{% endif %}{% if c.message %}; {{ c.message }}{% endif %}
{% else %}- none declared
{% endfor %}
Optional checks for this artifact (not required for release by the contract):
{% for c in review.optional %}- {{ c.name }}: {{ c.status }}{% if c.value is not none %}, normalised value {{ c.value }}{% endif %}{% if c.message %}; {{ c.message }}{% endif %}
{% else %}- none declared
{% endfor %}{% if review.other %}
Other recorded checks on this version:
{% for c in review.other %}- {{ c.name }}: {{ c.status }}{{ " (mandatory)" if c.mandatory else "" }}{% if c.value is not none %}, normalised value {{ c.value }}{% endif %}{% if c.message %}; {{ c.message }}{% endif %}
{% endfor %}{% endif %}
Lineage (scientific-input ancestors):
{% for a in obs.lineage %}- {{ a.logical_id }}: produced by {{ a.produced_by }}; checks {{ a.n_pass }} pass / {{ a.n_fail }} fail / {{ a.n_unknown }} unknown; resolved issues {{ a.resolved }}
{% else %}- none
{% endfor %}
## Budgets
Attempts on this artifact: {{ obs.attempts_used }} of {{ obs.max_attempts }}
Repairs remaining (episode): {{ obs.repairs_remaining }}
Cost remaining (episode): {{ obs.cost_remaining }}
\end{lstlisting}
\subsection*{Earlier reviewer, system prompt (pilot only)}
\begin{lstlisting}
You are the release reviewer inside a provenance-tracking scientific workflow controller.

A deliverable artifact has passed every blocking check and the controller proposes to accept it for release. You must decide whether the recorded evidence supports releasing it.

Rules:
- Use only the evidence shown: the contract, check results, open issues, the lineage summary, and the remaining budgets.
- approve: the evidence supports release.
- reject: the evidence indicates the artifact is likely wrong or not what the contract asks for.
- uncertain: the evidence is insufficient to decide.
- Anything other than approve blocks release of this version; the controller may then correct the artifact or withhold it.
- Reply with a single JSON object matching the schema: {"verdict": "approve" | "reject" | "uncertain", "reason": <one or two sentences>}. No other text.
\end{lstlisting}
\subsection*{Earlier reviewer, user prompt template (pilot only)}
\begin{lstlisting}
## Task contract
Workflow: {{ contract.workflow }}
Scientific target: {{ contract.scientific_target }}
Evidence tier: {{ contract.evidence_tier }}
Deliverables: {{ contract.deliverables | join(", ") }}
Declared tolerances:
{% for name, t in contract.tolerances %}- {{ name }}: abs {{ t.abs_tol }}, rel {{ t.rel_tol }}{% if t.unit %} ({{ t.unit }}){% endif %}
{% else %}- none
{% endfor %}Declared checks for this artifact:
{% for c in contract.node_checks %}- {{ c.name }} ({{ "mandatory" if c.mandatory else "optional" }}, cost {{ c.cost }})
{% else %}- none
{% endfor %}
## Release candidate
Artifact: {{ obs.logical_id }} (deliverable)
Produced by: {{ obs.produced_by }}{% if obs.last_outcome %}; last attempt outcome: {{ obs.last_outcome }}{% endif %}
Optional validators not yet run: {{ obs.optional_checks_available | join(", ") if obs.optional_checks_available else "none" }}

Check results on this version:
{% for c in obs.checks %}- {{ c.name }}: {{ c.outcome }}{{ " (mandatory)" if c.mandatory else "" }}{% if c.value is not none %}, normalised value {{ c.value }}{% endif %}{% if c.message %}; {{ c.message }}{% endif %}
{% else %}- none recorded
{% endfor %}
Lineage (scientific-input ancestors):
{% for a in obs.lineage %}- {{ a.logical_id }}: produced by {{ a.produced_by }}; checks {{ a.n_pass }} pass / {{ a.n_fail }} fail / {{ a.n_unknown }} unknown; resolved issues {{ a.resolved }}
{% else %}- none
{% endfor %}
## Budgets
Attempts on this artifact: {{ obs.attempts_used }} of {{ obs.max_attempts }}
Repairs remaining (episode): {{ obs.repairs_remaining }}
Cost remaining (episode): {{ obs.cost_remaining }}
\end{lstlisting}